\documentclass[11pt]{article} 

\usepackage[margin=1in]{geometry}
\usepackage{lipsum}       
\usepackage{graphicx}     
\usepackage{amsmath,amsthm}      
\usepackage{hyperref}     
\usepackage{natbib}       
\usepackage{balance}      

\usepackage{algorithm2e}
\usepackage{amsfonts}
\usepackage{tikz}
\newcommand{\real}{\mathbf{R}}

\usepackage{hyperref}
\usepackage{bbm}
\usepackage{algorithm2e}
\usepackage{amsfonts}
\usepackage{tikz}
\usepackage{multirow}
\usepackage{tcolorbox} 

\newcounter{mybox} 
\renewcommand{\themybox}{\arabic{mybox}} 

\usepackage{url}
\usepackage{xcolor}
\usepackage{color-edits}
\addauthor[Aleksandra]{ak}{green}

\addauthor[Kamesh]{km}{blue}

\addauthor[Sahasrajit]{sss}{orange}

\newtheorem{definition}{\textbf{Definition}}

\addauthor[Ashish]{ashish}{yellow}

\newtheorem{proposition}{Proposition}
\newtheorem{theorem}{Theorem}

\DeclareMathOperator*{\argmax}{arg\,max}

\makeatletter
\renewcommand\subsubsection{\@startsection{subsubsection}{3}{\z@}%
  {-2.25ex\@plus -1ex \@minus -.2ex}%
  {1.5ex \@plus .2ex}%
  {\normalfont\normalsize\bfseries}}
\makeatother

\title{\textbf{Multi-Selection for Recommendation Systems}}
\author{
    \textbf{Sahasrajit Sarmasarkar} \\
    Stanford University \\
    \texttt{sahasras@stanford.edu}
    \and
    \textbf{Zhihao Jiang} \\
    Stanford University \\
    \texttt{faebdc@stanford.edu}
    \and
    \textbf{Ashish Goel} \\
    Stanford University\\
    \texttt{ashishg@stanford.edu}
    \and 
    \textbf{Aleksandra Korolova} \\
    Princeton University \\
    \texttt{korolova@princeton.edu}
        \and
    \textbf{Kamesh Munagala\thanks{Supported by NSF awards CCF-2113798 and IIS-2402823.}} \\
    Duke University \\
    \texttt{kamesh@cs.duke.edu}
}
\date{} 

\begin{document}
    
\maketitle
\begin{abstract}
        We present the construction of a multi-selection model proposed in \cite{goel2024differentialprivacymultipleselections} to answer differentially private queries in the context of recommendation systems. The server sends back multiple recommendations and a ``local model'' to the user, which the user can run locally on its device to select the item that best fits its private features. We study a setup where the server uses a deep neural network (trained on the Movielens 25M dataset \cite{movielens}) as the ground truth for movie recommendation. In the multi-selection paradigm, the average recommendation utility is approximately 97\% of the optimal utility (as determined by the ground truth neural network) while maintaining a local differential privacy guarantee with $\epsilon$ ranging around 1 with respect to feature vectors of neighboring users. This is in comparison to an average recommendation utility of 91\% in the non-multi-selection regime under the same constraints.
    \end{abstract}

\maketitle

    \section{Introduction}


    Recommendation systems often track users through methods such as cookies \cite{mayer2012third}, cross-device tracking \cite{brookman2017cross}, and behavioral analysis \cite{kosinski2013private} to deliver personalized suggestions, enhancing user experience. However, these practices can lead to significant privacy risks, including data exploitation \cite{barocas2014big}, re-identification threats \cite{narayanan2008robust}, and surveillance concerns \cite{lyon2014surveillance}. To address these issues, several privacy-preserving techniques have been proposed, including differential privacy \cite{mcsherry2009differentially}, federated learning \cite{ammad2019federated}, homomorphic encryption \cite{kim2016efficient}, privacy-preserving matrix factorization \cite{hua2015dual}, and K-anonymity \cite{polat2005privacy}. Despite their potential, these methods often face challenges such as reduced utility, computational complexity, and communication overhead. In this work, we explore a privacy-preserving recommendation system where user queries are protected using differential privacy within the local trust model \cite{bebensee2019local}, with a focus on balancing the trade-offs between utility and privacy.

    In the local trust model, user queries and user features are  changed from the original to preserve privacy (typically by adding noise), which can lead to less accurate results from the server. To mitigate this issue, \cite{goel2024differentialprivacymultipleselections} introduced the concept of multi-selection, where the server returns multiple results, allowing the user to select the most relevant one without disclosing its choice to the server. To aid the user in selecting the most relevant result, the server can also provide a model to the user; the user can then plug in its true features (without the noise) into the model to choose the best option among the supplied results. This selection process can be handled by a software intermediary, such as a client application running on the user's device, which acts as the user's privacy delegate. The concept of using a proxy or browser extension on a user's device to select advertisements, aimed at enhancing privacy was first introduced in privacy-preserving ad systems like Adnostic \cite{toubiana2010adnostic} and Privad \cite{guha2011privad}. This high level architecture is shown in Figure \ref{fig:arch1}. In this paper, we assume that the underlying application requires that a single recommendation be served to the user, though the framework extends naturally to the case where the user needs to be served multiple results.

    The multi-selection approach has been shown to achieve provably good privacy-quality trade-offs in simple settings. Specifically, if the user features lie on a one-dimensional line and the user can easily determine which of the results returned by the server is the best, then for the same privacy guarantee, returning $k$ carefully chosen results can reduce the inaccuracy by a factor of $O(1/k)$~\cite{goel2024differentialprivacymultipleselections}. The key questions we ask in this paper are:
    \begin{enumerate}
        \item How do we extend the multi-selection approach to more complex (and more realistic) settings, where the user features are  multi-dimensional and where the user needs help choosing a single result from the set returned by the server?

        \item Does the multi-selection approach offer a better trade-off between privacy and accuracy than simpler baseline approaches such as computing a single result from noisy user features?
    
    \end{enumerate}

    We answer these questions by conducting an empirical case study. We start with the well-studied MovieLens 25M dataset~\cite{movielens}. We then train a neural network on this dataset exactly as described in \cite{github-recom}; for the purpose of our evaluation, we treat this model as ground truth. At a high-level, this paper makes two main contributions, corresponding to the two key-questions we outlined above:

    \begin{enumerate}
        \item  For the server, we propose a posterior sampling algorithm $\mathcal{A}_{sat-realuser}$ over the training set to construct a list of ``look-alike" users, and a greedy sub-modular maximisation \cite{nemhauser1978analysis} approach to generate the list of results to return to the user. We also propose a local PCA model that the server can send to the user to aid the user in choosing the best result among the ones returned. These algorithms have been designed to apply to fairly general settings and have natural interpretations. The details are in Section~\ref{sec:server_actions}.

\item We then compare our suite of algorithms against several baselines, demonstrating that our algorithms achieve substantially better accuracy for the same privacy guarantee. Our empirical results also show that the multi-selection approach provides a good privacy-accuracy tradeoff (details in Section~\ref{sec:experimental-results}).

    \end{enumerate}

Our work serves as a proof-of-concept, demonstrating the potential of the multi-selection architecture in privacy-preserving recommendation systems. We believe that this architecture should be considered as one viable option within the design space for anyone developing such systems.


The multi-selection architecture, along with privacy definitions and dis-utility models, is discussed in Section \ref{sec:simarch}. A detailed explanation of the server's actions in selecting the top $k$ movies and constructing the local model $\mathfrak{m}$ is provided in Section \ref{sec:server_actions}. The model training process is outlined in Section \ref{sec:model_construction_goals}, followed by an interpretation of geographic differential privacy in relation to standard local differential privacy guarantees in Section \ref{sec:sim_results_and_privacy}. Lastly, the simulation study, presented in Section \ref{sec:simulation}, demonstrates the superior performance of the posterior sampling algorithm $\mathcal{A}_{sat-realuser}$.



\subsection{Related Work}


\subsubsection{Local Differential privacy (LDP)}
LDP is a widely studied method for ensuring privacy in the local trust model \cite{bebensee2019local}. However, due to the independent noise addition to each data point, LDP often results in low utility \cite{neera2021privateutilityenhancedrecommendations,privacyenhancedmatfact}. To address this, the bounded Laplace mechanism was proposed at the user level and a mixture of Gaussian models at the server level to enhance utility in \cite{neera2021privateutilityenhancedrecommendations}. Additionally, dimensionality reduction techniques and a binary mechanism based on sampling was suggested in \cite{privacyenhancedmatfact} to improve utility. While these approaches focus on the training of models with LDP data, our work focuses on making inferences from LDP queries on a trained machine learning model.




%


\subsubsection{Multi-Selection}
%
An architecture for multi-selection, particularly with the goal of privacy-preserving advertising, was already introduced in \textit{Adnostic} by~\cite{toubiana2010adnostic}. Their proposal was to have a browser extension that would run the targeting and ad selection on the user's behalf, reporting to the server only click information using cryptographic techniques. Similarly, \textit{Privad} by~\cite{guha2011privad} propose to use an anonymizing proxy that operates between the client that sends broad interest categories to the proxy and the advertising broker, that transmits all ads matching the broad categories, with the client making appropriate selections from those ads locally. Although both Adnostic and Privad reason about the privacy properties of their proposed systems, unlike our work, neither provides DP guarantees. In our work, we give geographic differential privacy guarantees on the user query and relate it to local differential privacy in a small neighbourhood.




The multi-selection problem was introduced and studied theoretically in \cite{goel2024differentialprivacymultipleselections} assuming that every point along a one dimensional line could be a valid user query. However this assumption may not always be realistic as it may not generalize for most machine learning models such as neural networks and random forests. Thus, we restrict ourselves to sample from the feature vectors of the training set $A^{tr}$ while constructing the set of $k$ results $B_i$ and the local model $\mathfrak{m}$ given that the training set of users is public. 



\subsubsection{Homomorphic encryption}
A very recent work in \cite{cryptoeprint:2023/1438} presents a private web browser which receives homomorphic encrypted queries from the user, the query includes the cluster center $i^{*}$ and the search text $q$. The server sends the cosine similarity of every document in the cluster $i^{*}$ with the search text $q$ and the user can choose the index of document with the most similarity. Finally to retrieve the url of the matching documents, private information retrieval \cite{pir} is used. This essentially requires making the whole set of cluster centers public and the user to identify the cluster center $i^{*}$ that it is closest to. Both of these approaches significantly differ from our multi-selection model. Further, using homomorphic encryption for machine learning models \cite{cryptoeprint:2022/1602,Chillotti2020} typically comes with challenges such as high computation time and low utility, thus preventing its practical deployment. 
The notion of privacy achieved by our multi-selection framework is weaker than the one guaranteed by homomorphic encryption; however, the multi-selection setting has the advantage of placing fewer demands on the recommendation service to make its data / index essentially public. We, therefore, believe both frameworks are valuable but different additions to the private recommendation system toolkit, with different trade-offs.







\section{Overview of Multi-Selection Architecture}{\label{sec:simarch}}


At a high level, our multi-selection system architecture is shown in Figure \ref{fig:arch1}.  The on-device software intermediary applies a possibly randomised algorithm $\mathcal{A}$ on the user's input which it sends as a signal to the server.  The server takes as input a privatized user signal and returns to the user a small set of results as well as a compressed ML model. Using this compressed ML model (or the local model), the on-device software intermediary of the user decides, unknown to the server, which of the server responses to select given access to the true user input (true query and user features). 

\definecolor{usr}{rgb}{0.7,0.85,1.0}
\definecolor{usrl}{rgb}{0.9,0.95,1.0}
\definecolor{usrline}{rgb}{0.8,0.9,1.0}
\definecolor{sev}{rgb}{1.0,0.85,0.7}
\definecolor{sevl}{rgb}{1.0,0.95,0.9}
\definecolor{sevline}{rgb}{1.0,0.9,0.8}

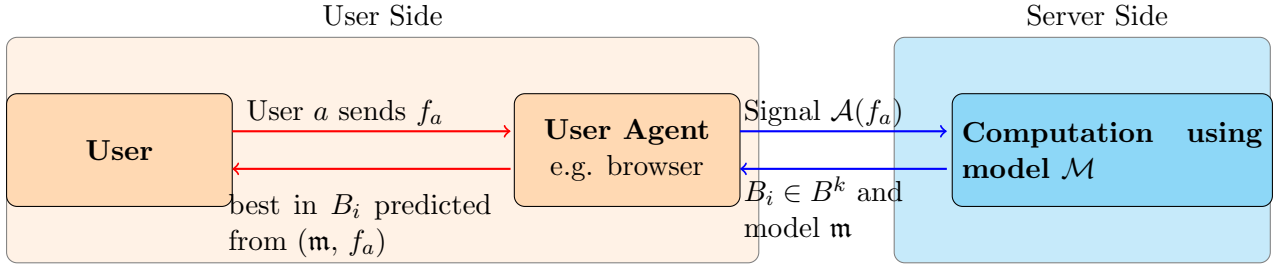
\begin{figure*}[htbp]
    \centering
    \tikzset{global scale/.style={
            scale=#1,
            every node/.append style={scale=#1}
        }
    }
    \begin{tikzpicture}
    
        \node[fill=sevl,draw=gray,rounded corners, minimum width=10cm, minimum height=3cm] at (2.5,2) {};
        
        \node at (2.5,3.8) {User Side};
    
        \node[fill=sev,draw=black,rounded corners, minimum width=3cm, minimum height=1.5cm] at (-1,2) {\bf User};
        
        \node[fill=sev,draw=black,rounded corners, minimum width=3cm, minimum height=1.5cm,align=center] at (5.75,2) {\bf User Agent \\ e.g. browser};
    
        \node[fill=cyan!20,draw=gray,rounded corners, minimum width=5cm, minimum height=3cm] at (11.8,2) {};
        
        \node at (12,3.8) {Server Side};
        \node[fill=cyan!40,draw=black,rounded corners, minimum width=4cm, minimum height=1.5cm] at (12.2,2) {\parbox{4 cm}{\bf Computation using model $\mathcal{M}$}};
    
        \draw[->,thick,red] (0.5,2.25) -- (4.2,2.25);         
        \draw[<-,thick,red] (0.5,1.75) -- (4.2,1.75);         
        \draw[->,thick,blue] (7.25,2.25) -- (10,2.25);         
        \draw[<-,thick,blue] (7.25,1.75) -- (10,1.75);
    
        \node at (2.0,2.5) {User $a$ sends $f_a$};
        \node[align=center] at (2.2,1.0) {\parbox{3.5 cm}{best in $B_i$ predicted from ($\mathfrak{m}$, $f_a$)}};
        \node at (8.55,2.5) {\parbox{2.5 cm}{Signal  $\mathcal{A}(f_a)$}};
        \node at (8.55,1.25) {\parbox{2.5cm}{$B_i \in B^k$ and \\ model $\mathfrak{m}$}};
    
    \end{tikzpicture}
    \caption{Overall architecture for multi-selection.}    
    \label{fig:arch1}
\end{figure*}

We now specify the various components of this architecture. We first present the notion of geographic differential privacy, and subsequently introduce the actions available to the user and server. We present the algorithmic components of the server actions in the next section.
    

\subsection{Geographic Differential Privacy}{\label{sec:privacy_definitions}}

We denote the set of users and results by $A$ and $B$ respectively. We represent the feature vector for user $a \in A$ as $f_a \in \mathbb{R}^d$. To keep things simple, we assume the feature vector includes the user query itself.  The server maintains a machine learning model $\mathcal{M}$ that takes as input a feature vector and returns a result or set of results. We assume that the set of  feature vectors on which the model $\mathcal{M}$ is trained is public. This assumption is standard; see~\cite{lowy2024optimaldifferentiallyprivatemodel,bu2024pretrainingdifferentiallyprivatemodels}. We denote this set by $A^{tr}$. The entire set $A$ is of course not public.  

    

We will use differential privacy as our notion of privacy of user features. This notion is  is introduced in \cite{dwork2006calibrating}; see \cite{dwork2014algorithmic} for a survey. We first define local differential privacy (LDP), which dates back to \cite{warner1965randomized}. This notion is standard, having been deployed by Google~\cite{RAPPOR} and Apple~\cite{Apple}. We refer the reader to Bebensee~\cite{bebensee2019local} for a survey. 

%
\begin{definition}[adapted from \cite{6686179,koufogiannis2015optimality}]
    Let $\epsilon>0$ be a desired level of privacy. Let $\mathcal{U}$ be a set of input data and $\mathcal{Y}$ be the set of all possible responses. Let $\Delta(\mathcal{Y})$ be the set of all probability distributions (over a sufficiently rich $\sigma$-algebra of $\mathcal{Y}$ given by $\sigma(\mathcal{Y})$). A mechanism $Q: \mathcal{U} \rightarrow \Delta(\mathcal{Y})$ is $\epsilon$-differentially private if for all $S \in \sigma(\mathcal{Y})$ and $u_1,u_2 \in \mathcal{U}$:
    $$ \mathbb{P}(Qu_1 \in S) \leq e^\epsilon \mathbb{P}(Qu_2\in S).$$
\end{definition}

In our context, it is unreasonable to insist a user is entirely indistinguishable from {\em all} other users -- the feature obfuscation needed to achieve this would render the results returned by the server to be hopelessly inaccurate. A more relevant notion of differential privacy in our context is  geographic differential privacy~\cite{geographicDP,alvim2018metricbasedlocaldifferentialprivacy} (GDP), which allows the privacy guarantee to decay with the distance between users. In other words, a user is indistinguishable from ``close by'' users in the feature space, while it may be possible to localize the user to a coarser region in space.  This notion has gained widespread adoption for anonymizing location data. In our context, it reflects, for instance, the intuition that the user is more interested in protecting the specifics of a medical query they are posing rather than protecting whether they are posing a medical query or an entertainment query.

Our use of geographic DP combines the definition in~\cite{geographicDP} with the trust assumptions of the local model, and is thus only a slight relaxation of the traditional local model. We restate the formal definition from \cite{koufogiannis2015optimality} and use it in the rest of this work.


%
\begin{definition}[adapted from \cite{koufogiannis2015optimality}]{\label{def:geo_DP}}
        Let $\epsilon>0$ be a desired level of privacy. Let $\mathcal{U}$ be a set of input data and $\mathcal{Y}$ be the set of all possible responses. Let $\Delta(\mathcal{Y})$ be the set of all probability distributions (over a sufficiently rich $\sigma$-algebra of $\mathcal{Y}$ given by $\sigma(\mathcal{Y})$). A mechanism $Q: \mathcal{U} \rightarrow \Delta(\mathcal{Y})$ is $\epsilon$-geographic differentially private if for all $S \in \sigma(\mathcal{Y})$ and $u_1,u_2 \in \mathcal{U}$:
    $$ \mathbb{P}(Qu_1 \in S) \leq e^{\epsilon|u_1-u_2|} \mathbb{P}(Qu_2\in S).$$
\end{definition}

One may observe that $\epsilon$-geographic differential privacy with respect to input data set $\mathcal{U}$ implies $\epsilon R$-local differential privacy with respect to user data set $\mathcal{U}'$ with diameter $R$ {\em i.e.}, where any two users $u_1,u_2 \in \mathcal{U}'$ satisfy $|u_1-u_2| \leq R$. A popular mechanism to satisfy geographic differential privacy is to add Laplace noise \cite{geographicDP}. We present details in Section \ref{sec:simarch}.

\subsection{Architecture Details} \label{sec:simarch}

We now instantiate each box in Figure \ref{fig:arch1}, motivate the choices made, and argue why it satisfies the desired privacy guarantees.

\paragraph{User Agent's action $\mathcal{A}$}: The agent independently adds noise to each component of the feature vector \( f_a \), with the noise being sampled from a Laplace distribution with parameter \( \eta \). We denote the corresponding high-dimensional Laplace distribution, centered at \( f_a \), as \( \mathcal{L}_{\eta}(f_a) \). This mechanism satisfies geographic differential privacy, as shown below. 



\begin{proposition}[adapted from \cite{koufogiannis2015optimality}]
    Let $s: \mathcal{U} \times \mathcal{Y} \rightarrow \mathbbm{R}$ by $L$-Lipschitz in $\mathcal{U}$. Then the mechanism $Q$ with density 
    $$ \mathbb{P}(Qu=y | u)\propto e^{s(u,y)} $$ is $\epsilon L$ geographic differentially private.
\end{proposition}

Choosing $s(u,y) = - \frac{|u-y|_1}{\eta}$ with $\mathcal{U}=\mathcal{Y}=\real^{d}$ now shows that the our noise addition mechanism $\mathcal{A}$ is ${1}/{\eta}$-geographic differentially private. Here, the distance in Definition \ref{def:geo_DP} is measured with respect to $\ell_1$ norm. In Section \ref{sec:user_action_privacy_deepnn}, we explore two interpretations of our noise addition mechanism within the framework of local differential privacy, specifically applied to the MovieLens 25M dataset~\cite{movielens}. 


\paragraph{Server's action.} The server's action can be split into three parts as described below in Box \ref{box:algorithm_overview}. We give a brief overview of each of the parts in this section, and defer the details to the next section.

\begin{tcolorbox}[title=Box \refstepcounter{mybox}\label{box:algorithm_overview}\themybox: Server's actions, colframe=black, colback=white, boxrule=0.5mm, width=1.0 \columnwidth, sharp corners, left=0mm, right=0mm, boxsep=0mm]
\begin{enumerate}
    \item \noindent Posterior sampling on receiving signal $f \in \real^d$.\\
\item \noindent Submodular maximization to select $k$ results.\\
\item \noindent Construction of a frugal model $\mathfrak{m}$.
\end{enumerate}


\end{tcolorbox}




(1) Given the privatized user feature vector $f$, the server attempts to maintains a prior over user features and update it to a posterior via Bayes rule.  However, this is not quite straightforward, since the server does not know the space of all users, and is hence unable to maintain a prior over it. We therefore refrain from computing the posterior distribution and work with a suitable guess of the posterior.  Denote this posterior by $\mathcal{D}$. We discuss the details of our posterior sampling algorithm $\mathcal{A}_{sat-realuser}$ in the next section. We present other candidate  sampling algorithms in Appendix \ref{sec:candidate_multi_selection}. 


(2) To select the set of $k$ results $B_i$, the server samples $q_1$ vectors from the posterior $\mathcal{D}$ computed in Step (1). It then greedily selects the $k$ results that optimize some sub-modular utility function $u_{(s)}(.)$. We discuss this in detail in the next section.

(3) In addition to returning the $k$ results, the server returns a compressed (frugal) model $\mathfrak{m}$ to the user to enable it to evaluate the quality of these results on the true feature vector. To construct $\mathfrak{m}$, the server samples $q_2$ feature vectors from the posterior  $\mathcal{D}$ and builds a PCA model on these samples. This simple model will therefore approximate the more complex model $\mathcal{M}$ within the neighborhood of the received signal $f$. This approach is inspired by LIME~\cite{10.1145/2939672.2939778}, which generates explanations by fitting a locally linear model through sampled points around an input. We provide a detailed construction in the next section.

\subsection{Measuring Utility of Results}{\label{sec:disutility}}
Recall that our main goal is to study the trade-off between privacy of the user features, and the quality of the returned results. Clearly, adding more noise leads to more privacy, but if the server has little clue what the true features are, the returned results will be inaccurate. Our framework mitigates this inaccuracy via choosing $k$ results, and we seek to study the tradeoff between $k$, the privacy parameter $\eta$, and the accuracy of the results.

There are two components to disutility of the result -- the loss due to privacy preserving noise, and the loss due to the user's software using a frugal model instead of $\mathcal{M}$ in evaluating results.  We  assume the loss (or utility) computed by the machine learning model \( \mathcal{M} \) is the ground truth. 

For the first component, we assume the user takes the set of results $B_i$ returned by the server and feeds them to $\mathcal{M}$ along with its true feature vector to find the result that yields highest utility. Thus, the dis-utility $d_i$ of an user $a$ from the sent of results $B_i$ sent by the server is given by 
\begin{equation}{\label{eq:disutility_inter}}
    d_i (a, B_i) = \max_{b \in B} u_{\mathcal{M}}(f_a,b) - \max\limits_{\substack{b': b' \in B_i}} u_{\mathcal{M}}(f_a,b')
\end{equation}

For the sum of the first two components, we assume the user uses the frugal model $\mathfrak{m}$ to choose the best result $b_f := \mathfrak{m}(f_a,B_i)$. The dis-utility $d_f$ of an user $a \in A$ from the result $b_f$ is given by 
\begin{equation}{\label{eq:disutility_final}}
    d_f (a, b_f) = \max_{b \in B} u_{\mathcal{M}}(f_a,b) - u_{\mathcal{M}}(f_a,b_f)
\end{equation}

\section{Instantiating server actions}{\label{sec:server_actions}} 

In this section, we first present the posterior distribution computed by the server given the signal $f$ sent by the user. We then present the algorithm that returns $k$ results to the user by sampling this posterior. We finally present the details of the frugal model $\mathfrak{m}$ sent back to the user, which enables the user to compute its best result.  

\subsection{Posterior Distribution}
We now define a posterior distribution $\mathcal{L}^{realuser}_{\eta}(f)$ below that will be used in the posterior sampling algorithm. 

\textbf{Distribution $\mathcal{L}^{realuser}_{\eta}(f)$}: Recall that $A^{tr}$ is the training set of users. We will use the term ``user'' and ``feature vector'' interchangably. For every user $a \in A^{tr}$, define distance $d_a = ||f - f_a||_1$. The posterior distribution $\mathcal{L}^{realuser}_{\eta}(f)$ samples a user $a \in A^{tr}$ with probability proportional to $\exp(-\frac{d_a}{\eta})$ and outputs its feature vector $f_a$. 

This distribution is identical to exponential mechanism \cite{exponentialmech} in differential privacy, however this distribution is now a function of signal $f$ that the user sends. 
    One may wonder why we restrict ourselves to sample from user feature vectors in the training set. We delve into this question in Appendix~\ref{sec:candidate_multi_selection}, where we present a multi-selection algorithm $\mathcal{A}_{sat}$ by defining the posterior over the entire feature space. We observe that such an algorithm attains higher dis-utility. We conjecture this is because the model $\mathcal{M}$ is trained over features corresponding to real users and not over the entire feature space. When noise is added to a real feature vector, the resulting feature vector may not map naturally to a real user, and the model output could have larger error.

\subsection{Greedy Result Selection} \label{sec:server_movie_selection}
\label{sec:posterior_sampling_alg}
We now present the algorithm $\mathcal{A}_{sat-realuser}$ used by the server to return the set of $k$ results. First, given the user signal, the server samples $q_1$ points in the user space $\mathbbm{R}^{d}$ from the posterior distribution $\mathcal{L}^{realuser}_{\eta}$. Call this set of sampled feature vectors as $F_s$. The server then defines a utility function $u_{(s)}(f,B)$. This function measures the utility of result set $B$ for a user with feature vector $f$. (Note that $f$ is now an arbitrary feature vector, and not the signal sent by the user.)  We define a general version of this utility where the user is interested in the top $t$ results instead of the top result.

\begin{equation}{\label{eq:utility_2_old}}
    u_{(s)}^{t}(f,B) = \max\limits_{B_c \subseteq B: |B_c| \leq {t}} \sum\limits_{b \in B_c} u_{\mathcal{M}}(f,b)
\end{equation}


Since the utility \( u_{(s)}(.) \) is evaluated based on the top \( t \) results in \( B_i \), we refer to this method as the posterior saturation algorithm. Further we refer this algorithm by ``realuser'' since it only samples from feature vectors of users in the training set. By setting \( t=1 \), this utility function aligns with the idea that the user chooses their best movie from the set of \( k \) results sent by the server.

The server now needs to compute the set $B$ of  that maximizes $U^t(B) = \sum_{f \in F_s} u^{t}_{(s)}(f,B)$. This function is a non-decreasing submodular function, where non-decreasing means $U(B) \geq U(A)$ for all $A \subseteq B$, and submodular means $U(A) + U(B) \geq U(A \cup B) + U(A \cap B)$ for all sets $A,B$. It is well-known that the greedy algorithm presented in Algorithm~\ref{alg:alg_gen} gives a $(1-\frac{1}{e})$ approximation of the optimal utility. 

\begin{theorem}[ \cite{nemhauser1978analysis}]\label{thm:greedy_selection}
    Consider a non-decreasing submodular function $U$ on the the subsets of a finite set $E$. Now consider the greedy algorithm that at each step chooses an element $i \in E\setminus S$ which maximises $U(S \cup \{i\})-U(S)$ and appends it to $S$. Then after $k$ steps,
    $$U(S) \geq \frac{e-1}{e} \max_{S^{*} \subseteq E; |S^{*}|=k} U(S^{*})$$
\end{theorem}

\RestyleAlgo{boxruled}
\LinesNumbered
\begin{algorithm}[ht]
\SetAlgoLined
\textbf{Parameters}: Distribution $\mathcal{P}$, utility $u_{(s)}(.,.)$.

Sample $q_1$ points in $\mathbb{R}^k$ from distribution $\mathcal{P}$ and call it $F_s$. 

Start with $B=\emptyset$.\\
\For{$\text{step} = 1,\ldots, k$}{
    Select result $b$ maximising $\sum\limits_{f\in F_s} u_{(s)}(f,\{B\cup \{b\}\})$.\label{maximal_expr}\\
    Update $B$ to $B \cup \{b\}$.
}
    \caption{Greedy Algorithm}
    \label{alg:alg_gen}
\end{algorithm}

\subsection{Construction of Frugal Model $\mathfrak{m}$}\label{sec:server_action_local_model}



We now discuss the construction of a frugal (or local) model $\mathfrak{m}$ that the server returns along with the results. 

Our goal is similar to the long line of work \cite{Fong_2017,Ribeiro_Singh_Guestrin_2018,10.1145/2939672.2939778} on making large machine learning models such as deep neural networks and random forests more interpretable in the neighbourhood of an input point. Typically, methods such as LIME~\cite{10.1145/2939672.2939778}  fit a locally linear model by sampling points in the neighbourhood of the input. Such methods have also been used to measure adversarial robustness~\cite{vora2023scoring,10.1145/3594869,10.1145/2939672.2939778}. 

Inspired by these works, we give the construction of a compressed PCA  model $\mathfrak{m}$  by sampling points from the posterior distribution $\mathcal{L}^{realuser}_{\eta}(f)$. Note that the algorithm below works for any posterior, and this will be important in our experiments, we consider other ways of constructing the posterior. 

\paragraph{PCA Algorithm.} The algorithm takes as input the result set $B$ and the posterior $\mathcal{L}^{realuser}_{\eta}(f)$. samples $q_2$ points from the posterior and forms a matrix $X\in \real^{q_2\times (1+d+k)}$, where the first element of each row is one, the next $d$ elements of each row contain sampled feature vector $f$, and the last $k$ elements denote the utility $u_{\mathcal{M}}(f,b)$ for every $b \in B$. We now consider SVD decomposition  $X= V\Sigma W^T$. For a parameter $p$, the server returns the top $p$ columns of $W$ to the user as the frugal model $\mathfrak{m}$, where these columns correspond to the top singular values in $\Sigma$. In our experiments, we choose $p = 20$.

\paragraph{User's action.} Let $W_L\in \real^{(1+d+k)\times p}$ be the first $p$ columns of $W$. The server sends $W_L$ along with the result set $B$ to the user $a\in A$. This user can find the best $x \in \real^{p}$ such that the first $1+d$ entries of $xW^T_L$ is closest to $[1; f_a]$ in $\ell_2$ norm. The remaining $k$ elements of vector $xW^T_L$ provide an estimation of $u_{\mathcal{M}}(f_a,b)$ for every $b \in B$. The user $a$ uses these values to choose the best result $b_f$ from $B$.



\section{ Dataset and model training } \label{sec:model_construction_goals}
We use the Movielens 25M dataset~\cite{movielens} to train a deep neural network to predict the rating a user assigns to a movie. The Movielens 25M dataset has $25,000,095$ ratings (on scale of 0-5) given by $162,541$ users for $62,423$ movies. Our training methodology and feature engineering is exactly the same as that described in~\cite{github-recom}.  We intentionally refrain from modifying the training methodology to ensure that the development of the multi-selection algorithm remains independent of the model. We give a brief description of the training methodology and feature engineering below.



\subsection{Training the neural network model}
We trained a deep neural network on randomly chosen subset of $250,000$ ratings from the first $500,000$ ratings, as outlined in \cite{github-recom}. This represents $10\%$ of the data. Among the users with these ratings, only those users who have rated a sufficient number of movies were used for training. As a result, our training set comprises of $3402$ users and nearly $17,000$ movies. Each user is represented by a $d=38$ dimensional vector and each movie is represented by a $19$ dimensional binary vector, which we describe later. The neural network takes a user and movie feature vector as an input and predicts the score on a scale of $0$ to $5$ that an user might assign to that movie. The trained neural network achieves a test accuracy of $61\%$ in predicting user-movie rating pairs up to an error of $0.5$ in the rating, with a test RMSE around $0.93$. 

For the multi-selection problem, given a user feature vector, the goal is to return a movie whose score predicted by the machine learning model is largest. For evaluating the multi-selection framework, we consider the entire set of $162,541$ users, since the feature vector of the user is private and has likely not been used for training.
 
\subsection{Feature Engineering}
\label{subsec:feature_engineering}
We describe the feature engineering  as done in~\cite{github-recom}. For each user, we generated a genre profile encompassing $19$ distinct genres. This profile included the number of movies each user likes (rating $\geq 4$) within each genre. To prevent bias towards users who have rated more movies, we scaled these values so that the sum of the scaled values across all genres is $1$. We do the same for disliked movies (rating $< 4$). Putting these together, we obtain $d=38$ user features.  The feature vector for each movie is also a $19$ dimensional binary vector denoting the genres this movie falls into. The features for every user and movie was constructed by iterating over the ratings in the entire dataset.

\section{Dataset Features and Privacy}{\label{sec:sim_results_and_privacy}}

In this section, we present some properties of the dataset, focusing particularly on interpreting geographic differential privacy in this context.

\subsection{Ratings of the Best Movie}
\begin{figure}[htbp]
 
  \centering
  \includegraphics[scale = 0.45]{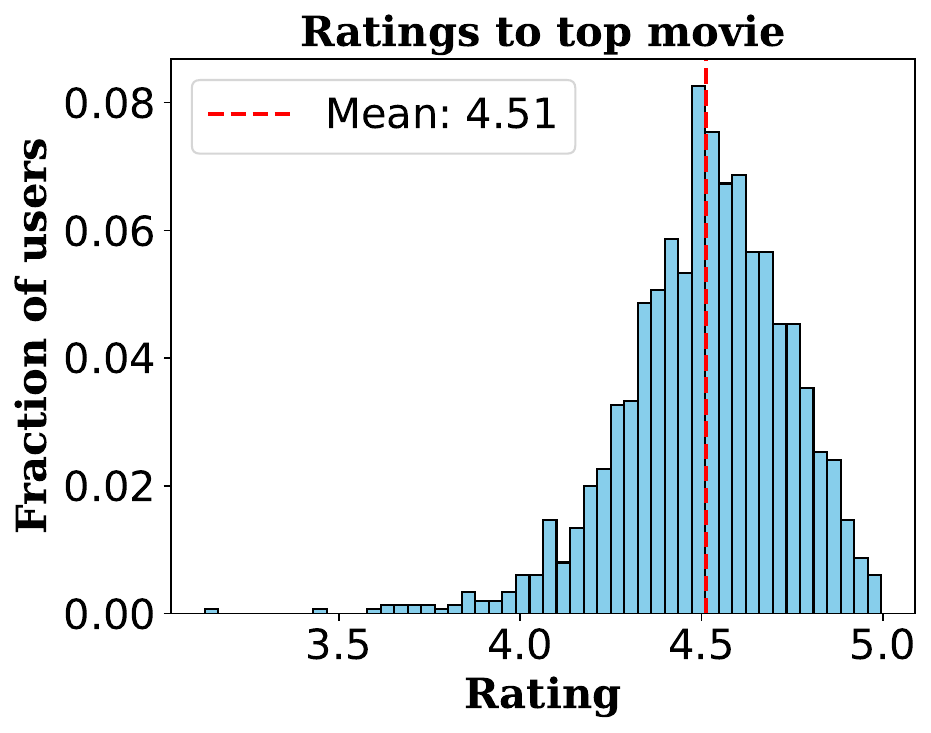}
\caption{Ratings predicted under Deep-NN model}
\label{fig:rating}
\end{figure}

{\label{sec:best_movie_ratings}}
In Figure \ref{fig:rating}, we first present the cumulative distribution of ratings that $1,500$ users, selected uniformly at random, assign to their most preferred movie according to the ground truth model \( \mathcal{M} \). This figure shows that the top-rated movie for a typical user generally receives a rating between 4 and 5, with an average rating of approximately $4.51$. This serves as a baseline for the user utility without any privacy guarantees, since this would be the result returned by the server had it known the true user features.

\subsection{Interpreting Geographic DP}
\label{sec:user_action_privacy_deepnn}
We next relate the geographic differential privacy guarantees (via Laplace noise addition of parameter $\eta$) to the more standard local differential privacy guarantee, albeit applied to a set of neighbouring feature vectors. The use of geographic DP is relatively new in our setting compared to local DP (where the privacy guarantees do not decay with distance). Hence, we believe that the results of this section will make it easier to interpret and motivate the Geographic DP guarantees used in our empirical evaluation, described later in Section~\ref{sec:experimental-results}.

Consider any two neighbouring users $a_1,a_2 \in A$ such that their feature vectors have a $\ell_1$ distance of at most $0.1$. Intuitively, these correspond to user feature vectors that differ on a couple of genres and agree on others. Choosing $\eta \in [0.05,0.2]$ in geographic DP achieves a local differential privacy guarantee of $0.1/ \eta \in [0.5,2]$ with respect to the feature vectors $\{f_{a_1},f_{a_2}\}$. 
Typically in local differential privacy \cite{bebensee2019local,RAPPOR,economicmethod}, a value of $\epsilon$ smaller than one is considered as a ``strong'' privacy guarantee and thus, we have a strong local privacy guarantee in the neighbourhood of an user. 

\begin{figure}[htbp]
  \centering
  \includegraphics[scale = 0.45]{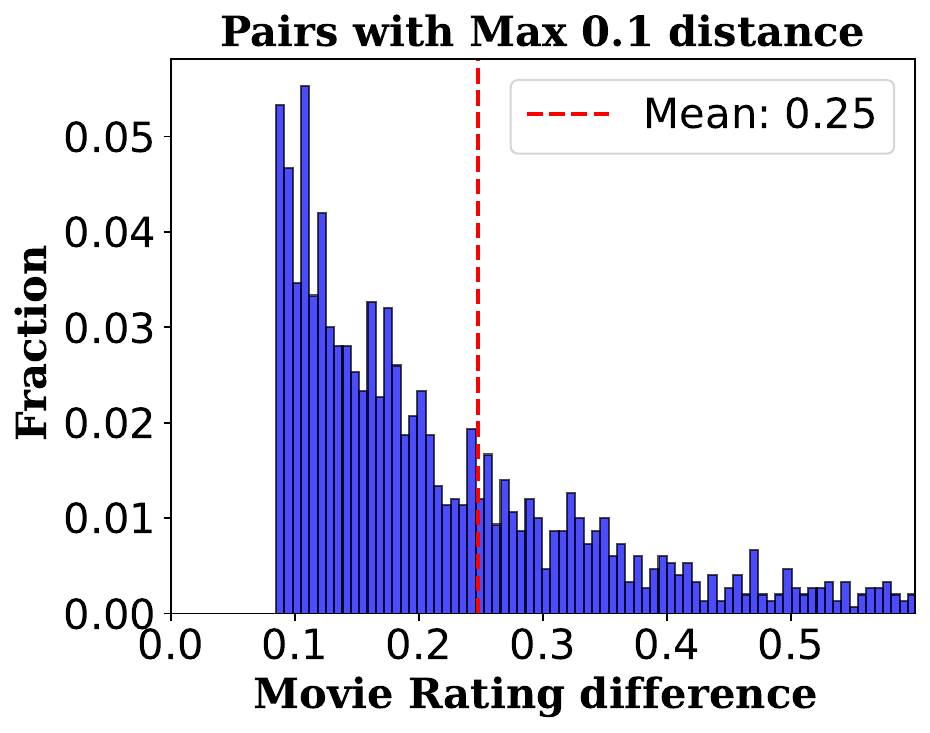}
\caption{Difference of mean ratings of neighboring users.}
\label{fig:rating_diff}
\end{figure}

We next ask whether users with such $\ell_1$ separation of $0.1$ are distinct enough. We show this is a few different ways.  We first compute the top $5$ movies preferred by each user based on the ground truth model \(\mathcal{M}\). We denote these sets as \(P_{a_1}\) and \(P_{a_2}\) for users \(a_1\) and \(a_2\), respectively. We then calculate the difference in the average ratings that user \(a_1\) assigns to the movies in \(P_{a_1}\) compared to those in \(P_{a_2}\). In Figure \ref{fig:rating_diff}, we plot the histogram  of these rating differences across $1500$ such user pairs \(\{a_1, a_2\}\), and observe a mean difference around 0.25, showing these users are sufficiently distinct. 

\begin{figure*}[htbp]
\centering
\begin{minipage}{0.33 \textwidth}
  \centering
  \includegraphics[scale = 0.35]{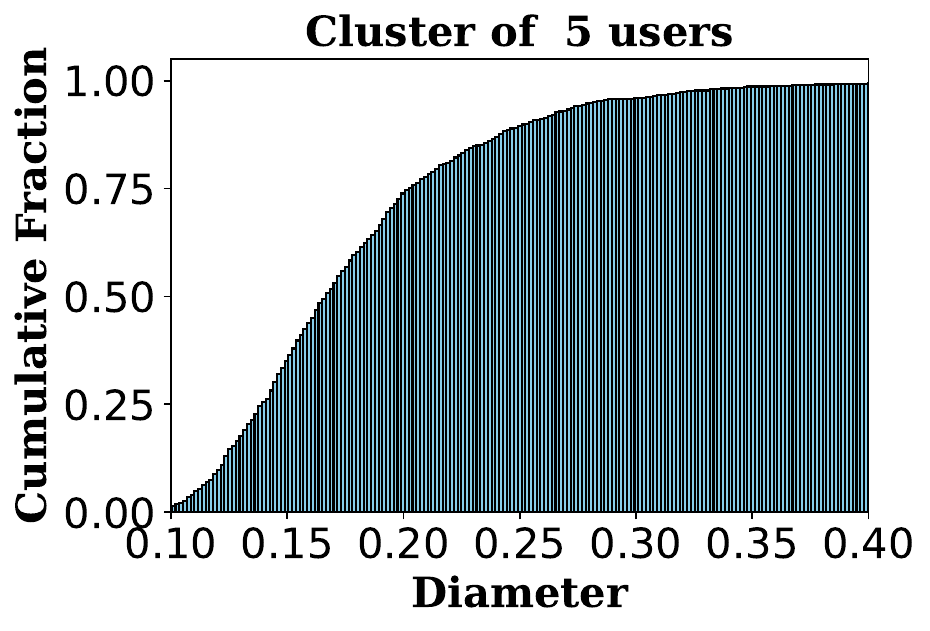}
\end{minipage}%
\begin{minipage}{.33\textwidth}
  \centering
  \includegraphics[scale = 0.35]{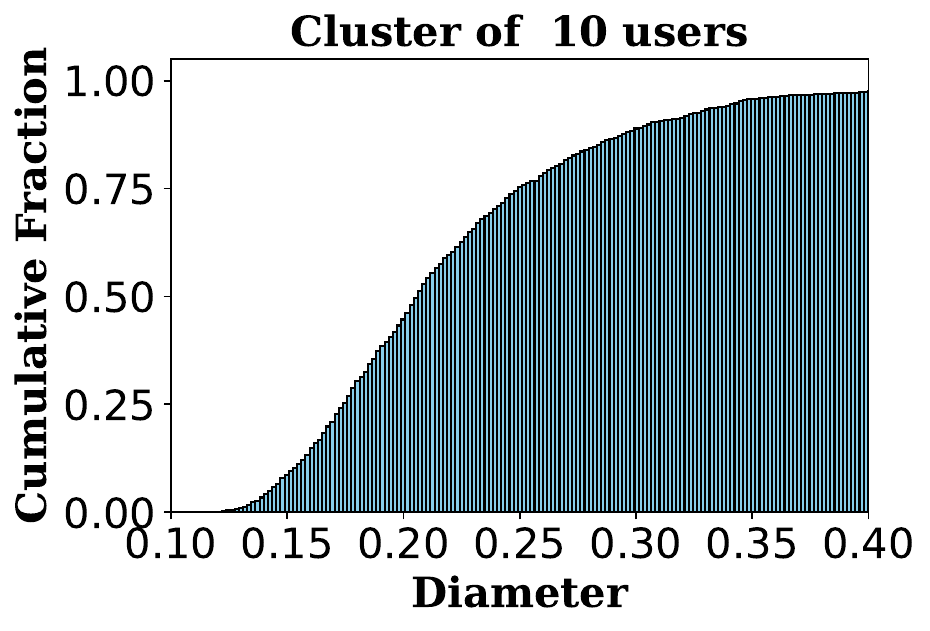}
\end{minipage}
\begin{minipage}{.33\textwidth}
  \centering
  \includegraphics[scale = 0.35]{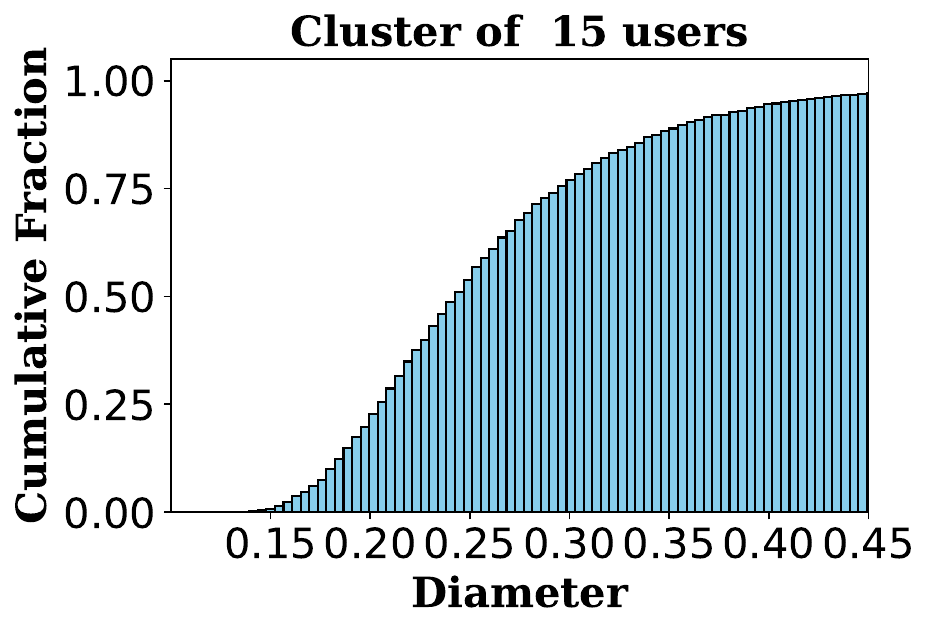}
\end{minipage}
\caption{Histogram plot of diameters of 5,10 and 15 sized clusters}
\label{fig:20}
\end{figure*}

\paragraph{User clusters.} Delving deeper, we sample $1500$ users from the set of users $A$ uniformly at random and build clusters of 5, 10 and 15 closest users centred around them. In Figure~\ref{fig:20}, we give a cumulative histogram plot of the cluster diameter of the corresponding feature vectors. The cluster diameter is defined as the largest $\ell_1$ distance of any two feature vectors of users in the cluster. 

\begin{figure}[htbp]
  \centering
  \includegraphics[scale = 0.45]{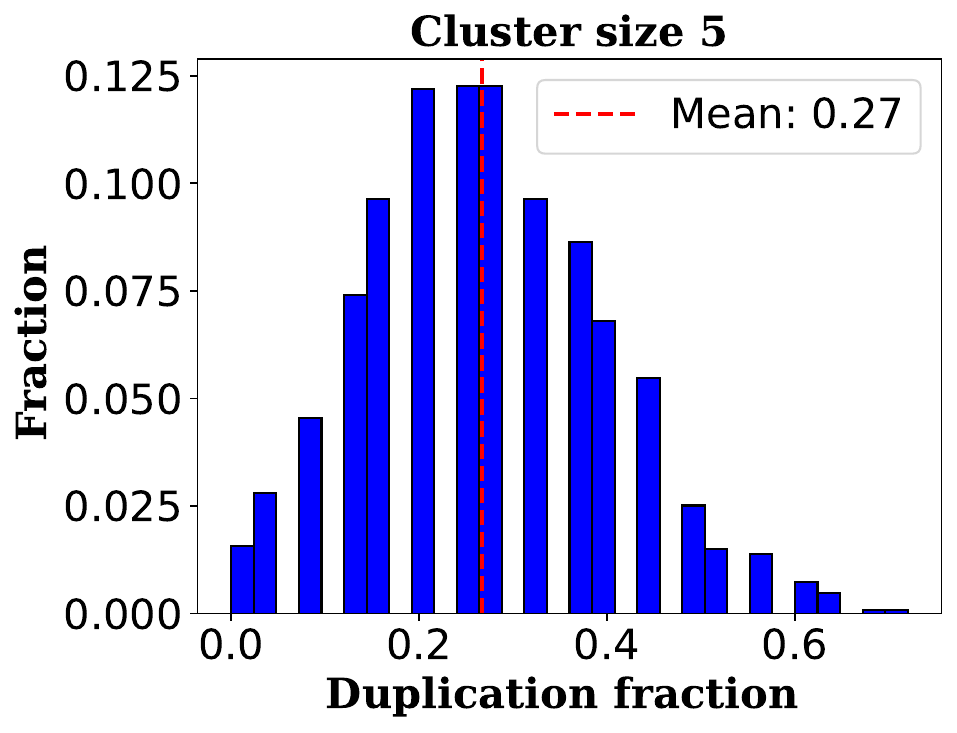}
\caption{Duplicated preferred movies of users in a cluster}
\label{fig:overlap_movies_cluster_5}
 
\end{figure}

\begin{table*}[htbp]
\begin{centering}
\begin{tabular}{|c|c|c|c|c|c|}
           \hline
                \multirow{2}{*}{{\centering $\eta$}} & \multirow{2}{*}{\parbox{2cm}{\centering $\mathcal{A}_{nopost} \newline (k=1)$}} & \multirow{2}{*}{\parbox{2.5cm}{\centering $\mathcal{A}_{nopost-realuser}\newline (k=1)$}} &\multicolumn{3}{|c|}{Values of $k$($\mathcal{A}_{sat-realuser}$)}  \\
                \cline{4-6} & & & 2 & 3 & 5 \\
                
\hline
0.03 & 0.1748 & 0.1591  & 0.0644 & 0.0441 & 0.0275\\
\hline
0.05 & 0.2622 & 0.1945  & 0.0811 & 0.0552 & 0.0354\\
\hline
0.1 & 0.3749 & 0.2826  & 0.1225 & 0.0849 & 0.0476  \\
\hline
0.15 & 0.3806 & 0.3548  & 0.1394 & 0.098 & 0.0577\\
\hline
0.2 & 0.4236 & 0.3897 & 0.1532 & 0.113 & 0.0673 \\
\hline
\end{tabular}
\caption{Dis-utility $d_i$  under $\mathcal{A}_{{sat-realuser}}$ compared to baselines}
\label{tab:no_local_model_disutilities_deepnn}
\end{centering}
\end{table*}

Observe that $\eta$-geographic DP for a cluster $C$ with diameter of $R$ implies $R/\eta$ local DP with respect to set of feature vectors $\{f_u: u \in C\}$. Observe from Figure \ref{fig:20} that nearly 75\%, 45\% and 24\% of clusters of size 5, 10 and 15 users respectively have a diameter at most $0.2$. Thus choosing an $\eta$ around $0.2$ enables us to achieve local DP guarantee with $\epsilon \approx 1$ with respect to the feature vectors of these clusters.

To determine if the preferred movies among users in these clusters are sufficiently different, we analyze the clusters of $5$ users. For each cluster $C$, we identify the 5 most preferred movies for each user within $C$. The multiset of these movies has size $25$. Suppose the size of the union of these sets is $q$. Then we use $1 - \frac{q}{25}$ as a measure of duplication. A smaller value implies more distinct sets. We plot this  histogram plot in Figure \ref{fig:overlap_movies_cluster_5}, observing a mean measure of duplication of 0.27. This shows the top movies for users in the cluster are sufficiently different. This shows local DP within the cluster goes a significant way towards preserving privacy.

\section{Simulation Study}
{\label{sec:simulation}}
We now study the trade-off between the number $k$ of returned results and the accuracy (or utility) for various choices of the privacy parameter $\eta$. We compare the algorithm presented before to several naive baselines, showing that our methodology yields significant improvement to the accuracy for small values of $k$. Further, we show the efficacy of the frugal model.

\subsection{Baseline Algorithms}
{\label{sec:naive_algorithm}}
Before proceeding further, we present several baseline server algorithms for posterior construction and result generation.  

\textbf{No-post algorithm $\mathcal{A}_{nopost}$}: In this case, the server sends back the top $k$ results for the signal $f$ received from the user. This is given by $\mathcal{R}^{k}_f = \argmax\limits_{S \subseteq B; |S|=k} \sum\limits_{b \in S}u_{\mathcal{M}}(f,b)$ 

\textbf{No-post (real-user) algorithm $\mathcal{A}_{nopost-realuser}$}: In the previous case, the signal $f$ need not correspond to a valid user. In this algorithm, the server 
finds the user $a$ in the training set $A^{tr}$ whose feature vector $f_a$ is the closest to received signal $f$. It then  
sends back the top $k$ results for the vector $f_a$. This is given by $\mathcal{R}^{k}_{f_a} = \argmax\limits_{S \subseteq B; |S|=k} \sum\limits_{b \in S}u_{\mathcal{M}}(f_a,b)$.

\textbf{Posterior ignore-signal algorithm $\mathcal{A}_{ig-sig}$}: The above baselines ignore posterior construction entirely. 
We now describe a baseline that intuitively captures the local trust applied to the {\em entire user space}, as opposed to the geographic DP model. In this baseline, the server ignores the signal from the server and just sends a set of $k$ results by sampling the users in its training set at random.

Formally, the server samples $q_1$ users uniformly at random from the training user set $A^{tr}$. Then it chooses a set of $k$ results to maximize the utility function $u_{(s)}(.)$ with respect to the sampled $q_1$ users. The utility function $u_{(s)}(.)$ is from algorithm $\mathcal{A}_{sat-realuser}$.




\subsection{Modifying $u^{(s)}(.)$ for Efficiency}
For computational efficiency, we only use the top $r$ results/movies for each user when calculating utility. We define the set of  top $r$ movies with the highest predicted ratings for a user with feature vector $f \in \mathbb{R}^d$ as $\mathcal{R}^r_f := \argmax\limits_{S \subseteq B; |S|=r} \sum\limits_{b \in S} u_{\mathcal{M}}(f, b)$. We will set $r = 100$ and appropriately define the utility function $u_{(s)}^{t,r}$ for algorithms $\mathcal{A}_{sat-realuser}$ and $\mathcal{A}_{ig-sig}$ as  below. One may observe that the this function continues to be sub-modular in $B$ for a given $f \in \real^d$.


\begin{equation}{\label{eq:utility_2}}
    u_{(s)}^{t,r}(f,B) = \max\limits_{B_c \subseteq B: |B_c| \leq {t}} \sum\limits_{b \in B_c} u_{\mathcal{M}}(f,b)\mathbbm{1}_{b \in \mathcal{R}^r_f}
\end{equation}


\subsection{Experimental Setup}{\label{sec:experimental_setup}}
In the previous section, we noted that selecting the noise parameter $\eta \in [0.05, 0.2]$ provides a good local differential privacy guarantee within a user's neighborhood. We therefore vary $\eta$ within this range.  In our experiments, we uniformly sample a user from the set of users \( A \), run the multi-selection framework, and repeat the experiment $1,500$ times to calculate the average dis-utility in the returned result set across the experiments.

We split the results into two parts. In the first part, we measure the dis-utility induced by geographic DP. In other words, the dis-utility is measured with respect to function $d_i$ (defined in equation \eqref{eq:disutility_inter}) assuming the user directly receives the $k$ movies $B$ from the server and it uses the ground truth model $\mathcal{M}$ to evaluate result quality. In the second part, we incorporate the dis-utility induced by the frugal model. In other words,  the dis-utility is measured with respect to function $d_f$ (defined in equation \eqref{eq:disutility_final}) where the agent (user's privacy delegate) uses the frugal PCA model $\mathfrak{m}$  to choose its movie $b_f$. 

\begin{figure*}[th]
\centering
\begin{minipage}{.5\textwidth}
  \centering
  \includegraphics[width=8cm, height=4cm]{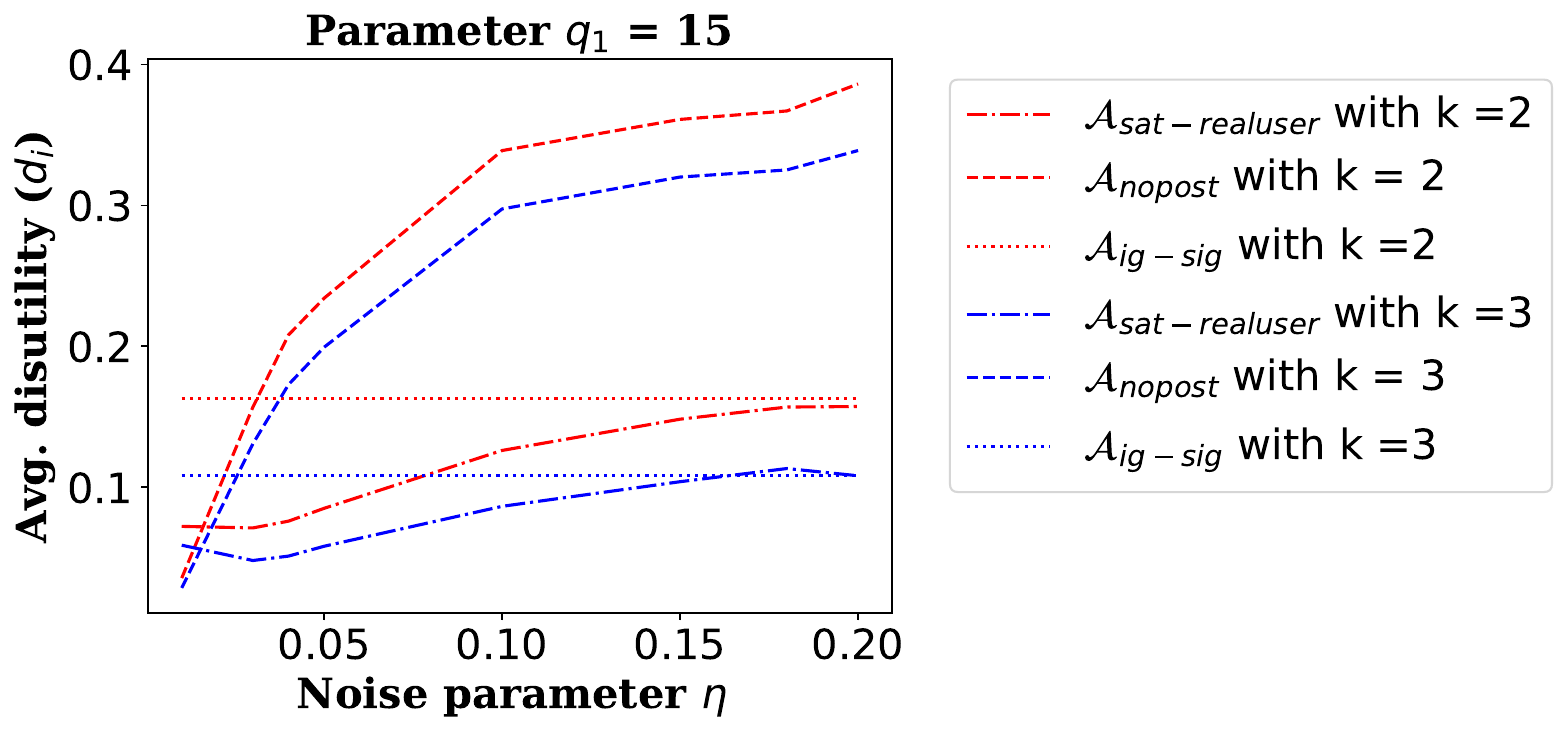}
 
\end{minipage}%
\begin{minipage}{.5\textwidth}
  \centering
  \includegraphics[width=8cm, height=4cm]{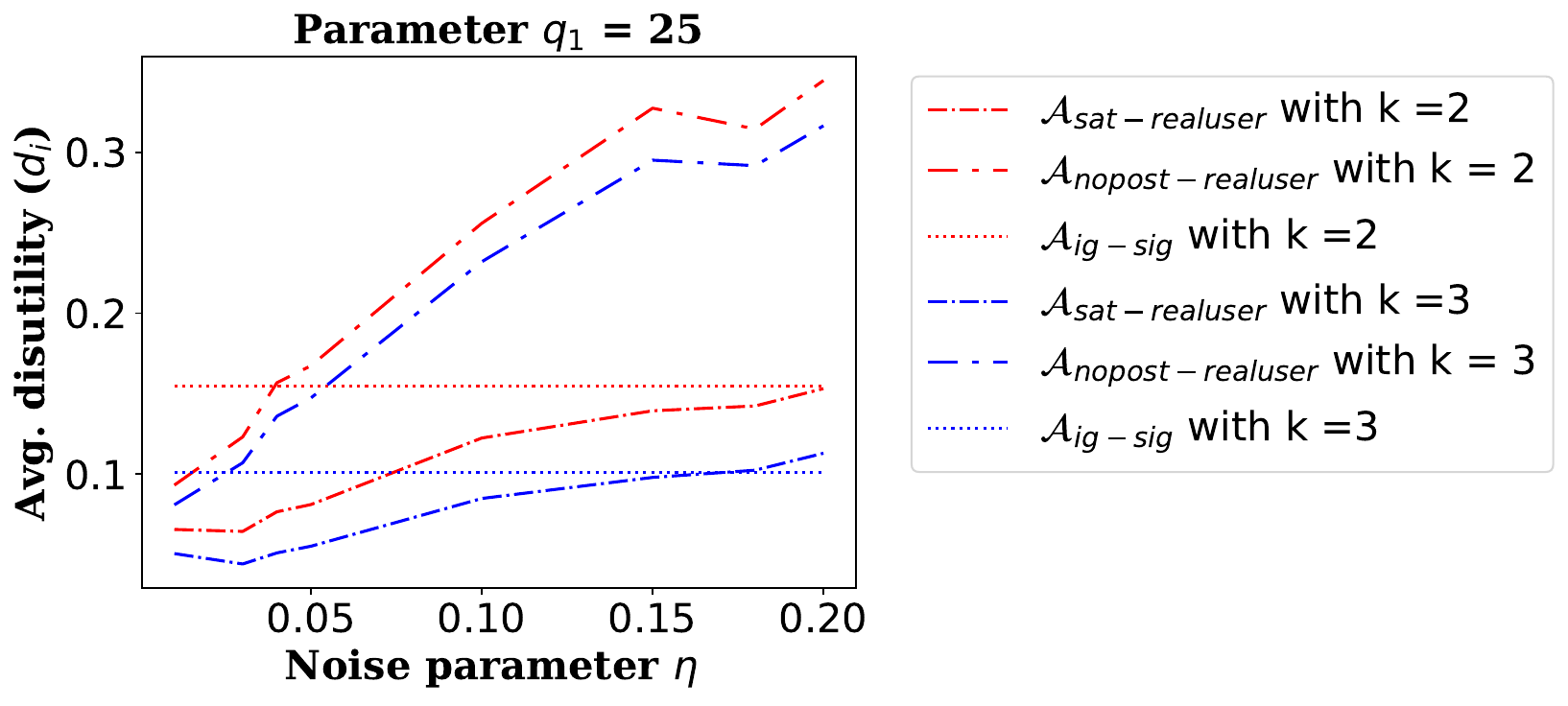}
  
\end{minipage}
\caption{Plotting dis-utilites $d_i$ of various different algorithms as a function of noise}
  \label{fig:diff_mech}
\end{figure*}

\begin{figure*}[t]
\centering
\begin{minipage}{.5\textwidth}
  \centering
  \includegraphics[width=8cm, height=4cm]{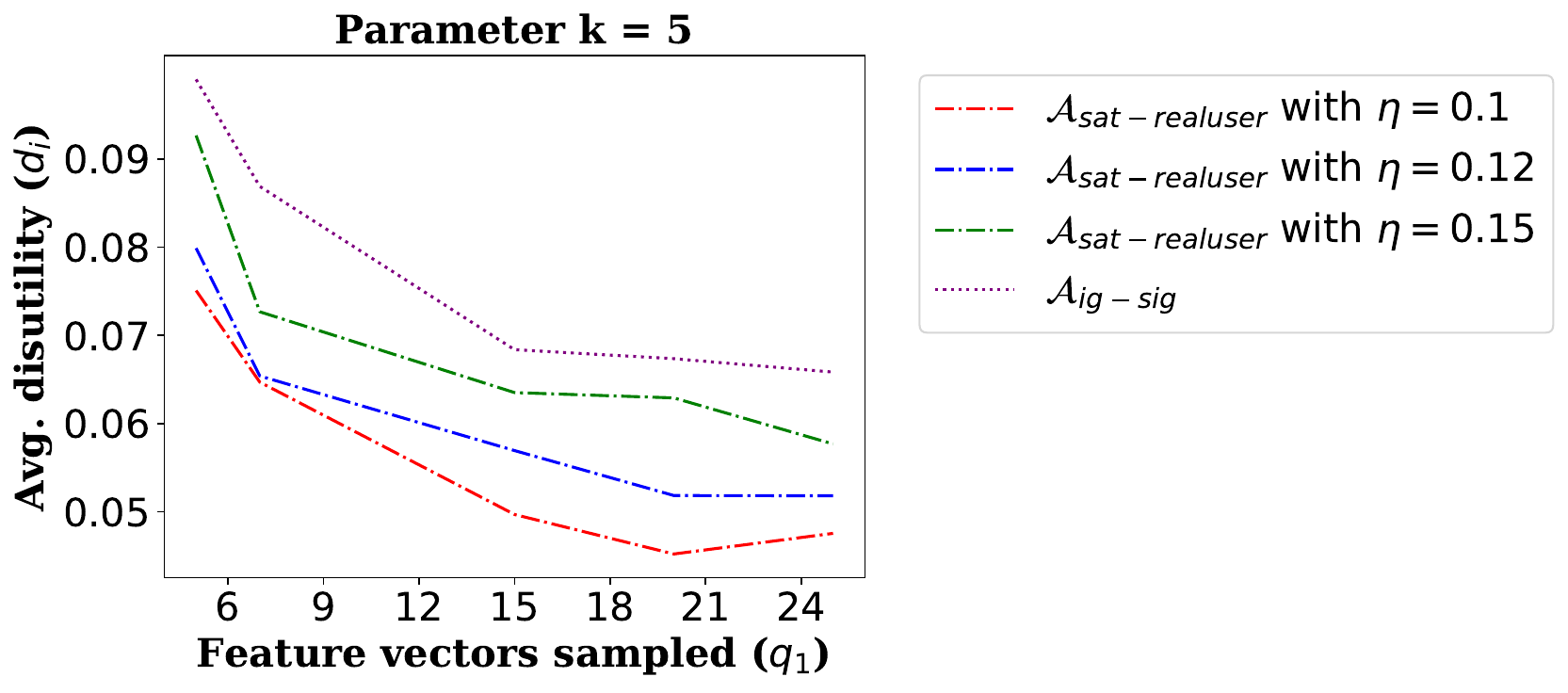}
    \caption{Dis-utility $d_i$ as a function of samples $q_1$}
     \label{fig:plot_q1}
\end{minipage}%
\begin{minipage}{.5\textwidth}
  \centering
  \includegraphics[width=8cm, height=4cm]{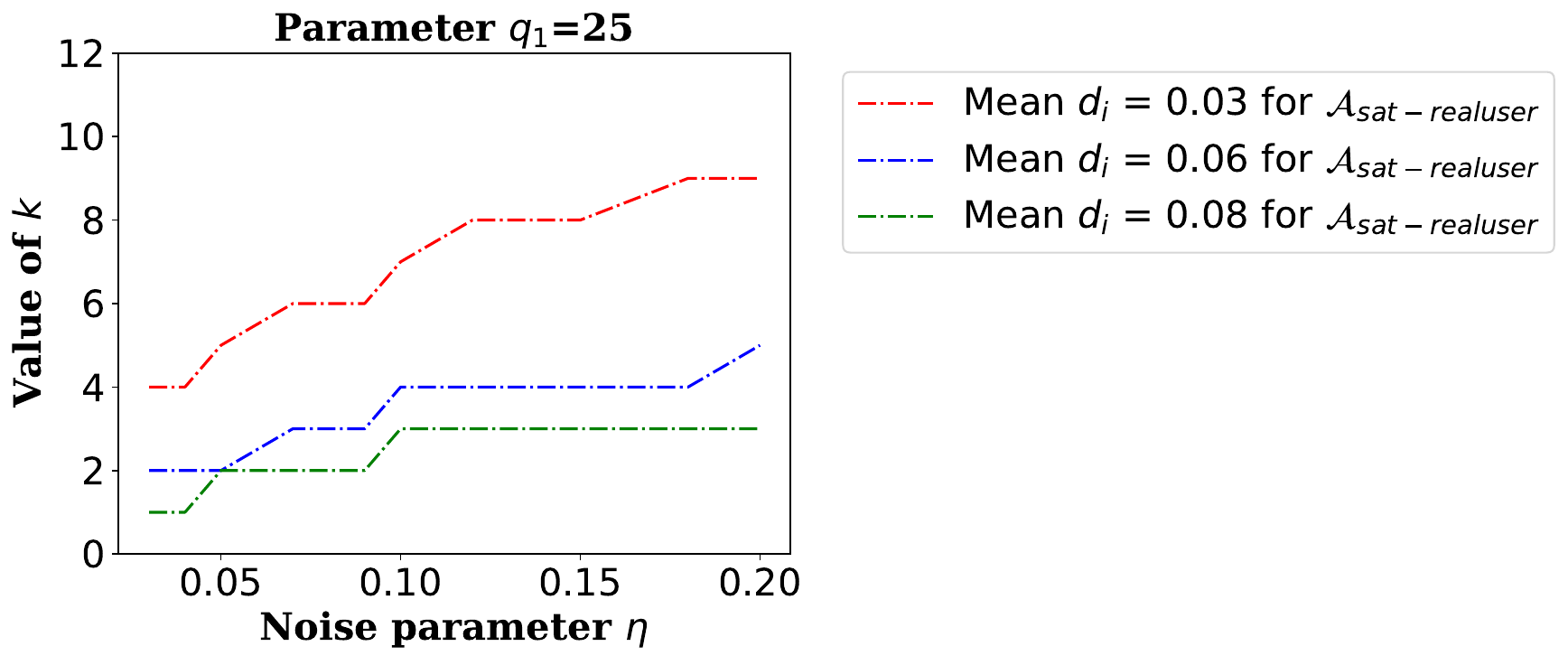}
  \caption{Variation of $k$ with $\eta$ for fixed dis-utility $d_i$}
 \label{fig:plot_with_k}
\end{minipage}%
\end{figure*}

\subsection{Experimental Results}
\label{sec:experimental-results}
At a high level, our experiments demonstrate the following.

\begin{itemize}
    \item The dis-utility monotonically decreases with $k$ and increases with noise level $\eta \in [0.05, 0.2]$.
    \item Our posterior sampling algorithm, \( \mathcal{A}_{sat-realuser} \), outperforms all baselines, with its dis-utility \( d_i \) decreasing monotonically as the number of samples \( q_1 \) from the posterior increases, stabilizing around $q_1 = 25$.
\end{itemize}

\begin{table*}[t]
\begin{centering}
\begin{tabular}{|c|c|c|c|c|c|c|c|c|}
           \hline
                \multirow{2}{*}{{\centering $\eta$}} & \multirow{2}{*}{\parbox{2cm}{\centering $\mathcal{A}_{nopost} \newline (k=1)$}} & \multirow{2}{*}{\parbox{2.5cm}{\centering $\mathcal{A}_{nopost-realuser}\newline (k=1)$}} &\multicolumn{3}{|c|}{Values of $k$($\mathcal{A}_{sat-realuser}$)} &\multicolumn{3}{|c|}{Values of $k$($\mathcal{A}_{ig-sig}$)} \\
                \cline{4-9} & & & 2 & 3 & 5 & 2 & 3 & 5 \\
              
                 \hline 0.03 & 0.194 & 0.159 & 0.111 & 0.121 & 0.121 & \multirow{5}{*}{\parbox{1 cm}{\centering 0.233}} & \multirow{5}{*}{\parbox{1 cm}{\centering 0.196}} & \multirow{5}{*}{\parbox{1 cm}{\centering  0.177}}  \\
 \cline{1-6}
0.05 & 0.268 & 0.194 & 0.132 & 0.117 & 0.12 & & &  \\
 \cline{1-6}
0.1 & 0.385 & 0.283 & 0.169 & 0.15 & 0.143 &  &  & \\
 \cline{1-6}
0.15 & 0.383 & 0.355 &  0.189 & 0.171 & 0.152 & & &  \\
 \cline{1-6}
0.2 & 0.423 & 0.39 & 0.198 & 0.173 & 0.167 &  &  & \\
 \hline

\end{tabular}
\caption{Dis-utility $d_f$ of our algorithm compared to baselines.}
\label{tab:local_model_disutilities_deepnn}
\end{centering}
\end{table*}

We thus demonstrate the idea of multi-selection holds promise for deep neural network based recommendation systems to answer differentially private queries.



\paragraph{Dis-utility due to Geographic DP}
\label{sec:model_access}
In this part, we measure the dis-utility as $d_i$, which assumes the user has access to $\mathcal{M}$. The results are shown in Table~\ref{tab:no_local_model_disutilities_deepnn}, where for different values of $\eta$, we show the accuracy of the framework improves with $k$, and improves over simple baselines.

In Figure \ref{fig:diff_mech}, we compare algorithm $\mathcal{A}_{{sat-realuser}}$ against baselines $\mathcal{A}_{nopost-realuser}$, $\mathcal{A}_{nopost}$ and $\mathcal{A}_{ig-sig}$ for different values of the samples $q_1$. One may observe that the mechanism $\mathcal{A}_{sat-realuser}$ gives the best dis-utility. Recall that the mechanism $\mathcal{A}_{ig-sig}$ ignores the signal itself and thus, its dis-utility is independent of the noise level $\eta$.


We next study the effect of the number of samples $q_1$. In Figure \ref{fig:plot_q1}, we compare the mechanisms $\mathcal{A}_{ig-sig}$ and $\mathcal{A}_{sat-realuser}$ for various values of $q_1$. We observe that the dis-utility monotonically decreases with $q_1$, and  saturates for $q_1=25$.  

In Figure~\ref{fig:plot_with_k}, we finally plot minimum value of $k$ needed to attain a fixed level of dis-utility under varying noise levels $\eta$ for $\mathcal{A}_{sat-realuser}$. We observe that even for stringent accuracy (low $d_i$) and privacy (high $\eta$) requirements, the value of $k$ is reasonable, being at most $10$.

\paragraph{Dis-utility due to Geographic DP and Frugal Model $\mathfrak{m}$}\label{sec:no_model_access}

We now assume the user selects the movie $b_f$ using the local model $\mathfrak{m}$. The  dis-utility is now given by $d_f$.  In Table~\ref{tab:local_model_disutilities_deepnn}, we show that the mechanism $\mathcal{A}_{{sat-realuser}}$ has far smaller dis-utility $d_f$ than the mechanisms $\mathcal{A}_{nopost}(k=1)$ and $\mathcal{A}_{nopost-realuser}(k=1)$. 

We do not compare the algorithms $\mathcal{A}_{nopost} (k>1)$ \\and $\mathcal{A}_{nopost-realuser} (k>1)$ in this part as these algorithms are not based on posterior sampling and thus are incompatible with the construction of a frugal model. We however did plot their dis-utilities in Figure \ref{fig:diff_mech} without invoking the frugal model, and observed that they perform much worse.

Thus, one may observe that the average empirical dis-utility $d_f$ is within 
$3 \%$ of the average utility of $4.51$ (from Figure \ref{fig:rating}) without any privacy guarantees. This means the average utility of our multi-selection approach is within $97\%$ of the optimal utility without privacy. However, the average utility of algorithms without multi-selection and posterior sampling is around $91\%$ of the optimal utility for noise level $\eta \approx 0.2$. This empirically demonstrates the benefits of multi-selection algorithms employing posterior sampling over naive baselines.

\section{Conclusion}

We present algorithmic innovations via posterior sampling and submodular optimization that make the framework for private recommendations proposed in \cite{goel2024differentialprivacymultipleselections} practical. We present a proof of concept study showing its promise for answering differentially private queries in a deep learning based movie recommendation model. It would be intriguing to investigate whether this approach can be extended to other commonly used machine learning models. Additionally, exploring alternative noise addition mechanisms beyond Laplace noise to enhance privacy preservation could also be of significant interest.

\bibliographystyle{plainnat} 
\bibliography{refs} 

\begin{thebibliography}{42}
\providecommand{\natexlab}[1]{#1}
\providecommand{\url}[1]{\texttt{#1}}
\expandafter\ifx\csname urlstyle\endcsname\relax
  \providecommand{\doi}[1]{doi: #1}\else
  \providecommand{\doi}{doi: \begingroup \urlstyle{rm}\Url}\fi

\bibitem[Alvim et~al.(2018)Alvim, Chatzikokolakis, Palamidessi, and Pazii]{alvim2018metricbasedlocaldifferentialprivacy}
Mário~S. Alvim, Konstantinos Chatzikokolakis, Catuscia Palamidessi, and Anna Pazii.
\newblock Metric-based local differential privacy for statistical applications, 2018.
\newblock URL \url{https://arxiv.org/abs/1805.01456}.

\bibitem[Ammad-Ud-Din et~al.(2019)Ammad-Ud-Din, Ivannikova, Khan, and Geifman]{ammad2019federated}
Muhammad Ammad-Ud-Din, Elena Ivannikova, Shahid~Ali Khan, and Dor Geifman.
\newblock Federated collaborative filtering for privacy-preserving personalized recommendation system.
\newblock \emph{arXiv preprint arXiv:1901.09888}, 2019.

\bibitem[Andr\'{e}s et~al.(2013)Andr\'{e}s, Bordenabe, Chatzikokolakis, and Palamidessi]{geographicDP}
Miguel~E. Andr\'{e}s, Nicol\'{a}s~E. Bordenabe, Konstantinos Chatzikokolakis, and Catuscia Palamidessi.
\newblock Geo-indistinguishability: Differential privacy for location-based systems.
\newblock In \emph{Proceedings of the 2013 ACM SIGSAC Conference on Computer \& Communications Security}, CCS '13, page 901–914, New York, NY, USA, 2013. Association for Computing Machinery.
\newblock ISBN 9781450324779.
\newblock \doi{10.1145/2508859.2516735}.
\newblock URL \url{https://doi.org/10.1145/2508859.2516735}.

\bibitem[Apple(2017)]{Apple}
Apple.
\newblock Learning with privacy at scale.
\newblock \emph{Apple Machine Learning Journal}, 1, 2017.
\newblock \url{https://machinelearning.apple.com/2017/12/06/learning-with-privacy-at-scale.html}.

\bibitem[Barocas and Nissenbaum(2014)]{barocas2014big}
Solon Barocas and Helen Nissenbaum.
\newblock Big data's end run around anonymity and consent.
\newblock In \emph{Privacy, Big Data, and the Public Good: Frameworks for Engagement}, pages 44--75. Cambridge University Press, 2014.

\bibitem[Bebensee(2019)]{bebensee2019local}
Bj{\"o}rn Bebensee.
\newblock Local differential privacy: a tutorial.
\newblock \emph{arXiv preprint arXiv:1907.11908}, 2019.

\bibitem[Brookman et~al.(2017)Brookman, Rouge, Alva, and Yeung]{brookman2017cross}
Justin Brookman, Pierre Rouge, Akua Alva, and Christo~Wilson Yeung.
\newblock Cross-device tracking: Measurement and disclosures.
\newblock In \emph{Proceedings of the Privacy Enhancing Technologies Symposium}, 2017.

\bibitem[Bu et~al.(2024)Bu, Zhang, Hong, Zha, and Karypis]{bu2024pretrainingdifferentiallyprivatemodels}
Zhiqi Bu, Xinwei Zhang, Mingyi Hong, Sheng Zha, and George Karypis.
\newblock Pre-training differentially private models with limited public data, 2024.
\newblock URL \url{https://arxiv.org/abs/2402.18752}.

\bibitem[Chillotti et~al.(2020)Chillotti, Joye, and Paillier]{Chillotti2020}
Ilaria Chillotti, Marc Joye, and Pascal Paillier.
\newblock New challenges for fully homomorphic encryption.
\newblock In \emph{Privacy Preserving Machine Learning - PriML and PPML Joint Edition Workshop, NeurIPS 2020}, December 2020.
\newblock URL \url{https://neurips.cc/virtual/2020/19640}.

\bibitem[Chor et~al.(1995)Chor, Goldreich, Kushilevitz, and Sudan]{pir}
B.~Chor, O.~Goldreich, E.~Kushilevitz, and M.~Sudan.
\newblock Private information retrieval.
\newblock In \emph{Proceedings of the 36th Annual Symposium on Foundations of Computer Science}, FOCS '95, page~41, USA, 1995. IEEE Computer Society.
\newblock ISBN 0818671831.

\bibitem[Dao-V(2024)]{github-recom}
Dao-V.
\newblock Movie recommendation system, 2024.
\newblock URL \url{https://github.com/dao-v/Movie_Recommendation_System}.
\newblock Accessed: 2024-08-11.

\bibitem[Duchi et~al.(2013)Duchi, Jordan, and Wainwright]{6686179}
John~C. Duchi, Michael~I. Jordan, and Martin~J. Wainwright.
\newblock Local privacy and statistical minimax rates.
\newblock In \emph{2013 IEEE 54th Annual Symposium on Foundations of Computer Science}, pages 429--438, 2013.
\newblock \doi{10.1109/FOCS.2013.53}.

\bibitem[Dwork et~al.(2006)Dwork, {McSherry}, Nissim, and Smith]{dwork2006calibrating}
Cynthia Dwork, Frank {McSherry}, Kobbi Nissim, and Adam Smith.
\newblock Calibrating noise to sensitivity in private data analysis.
\newblock In \emph{Theory of Cryptography: Third Theory of Cryptography Conference, TCC 2006, New York, NY, USA, March 4-7, 2006. Proceedings 3}, pages 265--284. Springer, 2006.

\bibitem[Dwork et~al.(2014)Dwork, Roth, et~al.]{dwork2014algorithmic}
Cynthia Dwork, Aaron Roth, et~al.
\newblock The algorithmic foundations of differential privacy.
\newblock \emph{Foundations and Trends{\textregistered} in Theoretical Computer Science}, 9\penalty0 (3--4):\penalty0 211--407, 2014.

\bibitem[Erlingsson et~al.(2014)Erlingsson, Pihur, and Korolova]{RAPPOR}
\'{U}lfar Erlingsson, Vasyl Pihur, and Aleksandra Korolova.
\newblock {RAPPOR}: Randomized aggregatable privacy-preserving ordinal response.
\newblock In \emph{Proceedings of the 2014 ACM SIGSAC Conference on Computer and Communications Security}, CCS '14, pages 1054--1067, New York, NY, USA, 2014. ACM.
\newblock ISBN 978-1-4503-2957-6.
\newblock URL \url{http://doi.acm.org/10.1145/2660267.2660348}.

\bibitem[Fong and Vedaldi(2017)]{Fong_2017}
Ruth~C. Fong and Andrea Vedaldi.
\newblock Interpretable explanations of black boxes by meaningful perturbation.
\newblock In \emph{2017 IEEE International Conference on Computer Vision (ICCV)}. IEEE, October 2017.
\newblock \doi{10.1109/iccv.2017.371}.
\newblock URL \url{http://dx.doi.org/10.1109/ICCV.2017.371}.

\bibitem[Goel et~al.(2024)Goel, Jiang, Korolova, Munagala, and Sarmasarkar]{goel2024differentialprivacymultipleselections}
Ashish Goel, Zhihao Jiang, Aleksandra Korolova, Kamesh Munagala, and Sahasrajit Sarmasarkar.
\newblock Differential privacy with multiple selections, 2024.
\newblock URL \url{https://arxiv.org/abs/2407.14641}.

\bibitem[Guha et~al.(2011)Guha, Cheng, and Francis]{guha2011privad}
Saikat Guha, Bin Cheng, and Paul Francis.
\newblock Privad: Practical privacy in online advertising.
\newblock In \emph{USENIX conference on Networked systems design and implementation}, pages 169--182, 2011.

\bibitem[Han et~al.(2023)Han, Lin, Shen, Wang, and Guan]{10.1145/3594869}
Sicong Han, Chenhao Lin, Chao Shen, Qian Wang, and Xiaohong Guan.
\newblock Interpreting adversarial examples in deep learning: A review.
\newblock \emph{ACM Comput. Surv.}, 55\penalty0 (14s), jul 2023.
\newblock ISSN 0360-0300.
\newblock \doi{10.1145/3594869}.
\newblock URL \url{https://doi.org/10.1145/3594869}.

\bibitem[Harper and Konstan(2015)]{movielens}
F.~Maxwell Harper and Joseph~A. Konstan.
\newblock The movielens datasets: History and context.
\newblock \emph{ACM Trans. Interact. Intell. Syst.}, 5\penalty0 (4), dec 2015.
\newblock ISSN 2160-6455.
\newblock \doi{10.1145/2827872}.
\newblock URL \url{https://doi.org/10.1145/2827872}.

\bibitem[Henzinger et~al.(2023)Henzinger, Dauterman, Corrigan-Gibbs, and Zeldovich]{cryptoeprint:2023/1438}
Alexandra Henzinger, Emma Dauterman, Henry Corrigan-Gibbs, and Nickolai Zeldovich.
\newblock Private web search with tiptoe.
\newblock Cryptology ePrint Archive, Paper 2023/1438, 2023.
\newblock URL \url{https://eprint.iacr.org/2023/1438}.
\newblock \url{https://eprint.iacr.org/2023/1438}.

\bibitem[Hsu et~al.(2014)Hsu, Gaboardi, Haeberlen, Khanna, Narayan, Pierce, and Roth]{economicmethod}
J.~Hsu, M.~Gaboardi, A.~Haeberlen, S.~Khanna, A.~Narayan, B.~C. Pierce, and A.~Roth.
\newblock Differential privacy: An economic method for choosing epsilon.
\newblock In \emph{2014 IEEE 27th Computer Security Foundations Symposium (CSF)}, pages 398--410, Los Alamitos, CA, USA, jul 2014. IEEE Computer Society.
\newblock \doi{10.1109/CSF.2014.35}.
\newblock URL \url{https://doi.ieeecomputersociety.org/10.1109/CSF.2014.35}.

\bibitem[Hua and Xiong(2015)]{hua2015dual}
Jinfei Hua and Li~Xiong.
\newblock A dual mechanism for privacy-preserving data sharing with enhanced utility.
\newblock \emph{Data \& Knowledge Engineering}, 96:\penalty0 1--20, 2015.

\bibitem[Kim et~al.(2016)Kim, Song, Kim, Lee, and Lee]{kim2016efficient}
Min Kim, Young-Sik Song, Seung-Hoon Kim, Hwanjo Lee, and Jaesik Lee.
\newblock Efficient privacy-preserving collaborative filtering based on homomorphic encryption.
\newblock \emph{IEEE Transactions on Knowledge and Data Engineering}, 28\penalty0 (4):\penalty0 1004--1016, 2016.

\bibitem[Kosinski et~al.(2013)Kosinski, Stillwell, and Graepel]{kosinski2013private}
Michal Kosinski, David Stillwell, and Thore Graepel.
\newblock Private traits and attributes are predictable from digital records of human behavior.
\newblock \emph{Proceedings of the National Academy of Sciences}, 110\penalty0 (15):\penalty0 5802--5805, 2013.

\bibitem[Koufogiannis et~al.(2015)Koufogiannis, Han, and Pappas]{koufogiannis2015optimality}
Fragkiskos Koufogiannis, Shuo Han, and George~J Pappas.
\newblock Optimality of the laplace mechanism in differential privacy.
\newblock \emph{arXiv preprint arXiv:1504.00065}, 2015.

\bibitem[Lowy et~al.(2024)Lowy, Li, Huang, and Razaviyayn]{lowy2024optimaldifferentiallyprivatemodel}
Andrew Lowy, Zeman Li, Tianjian Huang, and Meisam Razaviyayn.
\newblock Optimal differentially private model training with public data, 2024.
\newblock URL \url{https://arxiv.org/abs/2306.15056}.

\bibitem[Lyon(2014)]{lyon2014surveillance}
David Lyon.
\newblock \emph{Surveillance, Privacy, and the Globalization of Personal Data}.
\newblock MIT Press, 2014.

\bibitem[Marcolla et~al.(2022)Marcolla, Sucasas, Manzano, Bassoli, Fitzek, and Aaraj]{cryptoeprint:2022/1602}
Chiara Marcolla, Victor Sucasas, Marc Manzano, Riccardo Bassoli, Frank~H.P. Fitzek, and Najwa Aaraj.
\newblock Survey on fully homomorphic encryption, theory, and applications.
\newblock Cryptology ePrint Archive, Paper 2022/1602, 2022.
\newblock URL \url{https://eprint.iacr.org/2022/1602}.
\newblock \url{https://eprint.iacr.org/2022/1602}.

\bibitem[Mayer and Mitchell(2012)]{mayer2012third}
Jonathan~R Mayer and John~C Mitchell.
\newblock Third-party web tracking: Policy and technology.
\newblock In \emph{2012 IEEE Symposium on Security and Privacy}, pages 413--427. IEEE, 2012.

\bibitem[McSherry and Mironov(2009)]{mcsherry2009differentially}
Frank McSherry and Ilya Mironov.
\newblock Differentially private recommender systems: Building privacy into the net.
\newblock In \emph{Proceedings of the 15th ACM SIGKDD International Conference on Knowledge Discovery and Data Mining}, pages 627--636. ACM, 2009.

\bibitem[McSherry and Talwar(2007)]{exponentialmech}
Frank McSherry and Kunal Talwar.
\newblock Mechanism design via differential privacy.
\newblock In \emph{48th Annual IEEE Symposium on Foundations of Computer Science (FOCS'07)}, pages 94--103, 2007.
\newblock \doi{10.1109/FOCS.2007.66}.

\bibitem[Narayanan and Shmatikov(2008)]{narayanan2008robust}
Arvind Narayanan and Vitaly Shmatikov.
\newblock Robust de-anonymization of large sparse datasets.
\newblock In \emph{2008 IEEE Symposium on Security and Privacy (sp 2008)}, pages 111--125. IEEE, 2008.

\bibitem[Neera et~al.(2021)Neera, Chen, Aslam, Wang, and Shu]{neera2021privateutilityenhancedrecommendations}
Jeyamohan Neera, Xiaomin Chen, Nauman Aslam, Kezhi Wang, and Zhan Shu.
\newblock Private and utility enhanced recommendations with local differential privacy and gaussian mixture model, 2021.
\newblock URL \url{https://arxiv.org/abs/2102.13453}.

\bibitem[Nemhauser et~al.(1978)Nemhauser, Wolsey, and Fisher]{nemhauser1978analysis}
George~L Nemhauser, Laurence~A Wolsey, and Marshall~L Fisher.
\newblock An analysis of approximations for maximizing submodular set functions—i.
\newblock \emph{Mathematical programming}, 14:\penalty0 265--294, 1978.

\bibitem[Polat and Du(2005)]{polat2005privacy}
Huseyin Polat and Wenliang Du.
\newblock Privacy-preserving collaborative filtering using randomized perturbation techniques.
\newblock In \emph{Proceedings of the 2005 IEEE International Conference on Data Mining}, pages 625--628. IEEE, 2005.

\bibitem[Ribeiro et~al.(2016)Ribeiro, Singh, and Guestrin]{10.1145/2939672.2939778}
Marco~Tulio Ribeiro, Sameer Singh, and Carlos Guestrin.
\newblock "why should i trust you?": Explaining the predictions of any classifier.
\newblock In \emph{Proceedings of the 22nd ACM SIGKDD International Conference on Knowledge Discovery and Data Mining}, KDD '16, page 1135–1144, New York, NY, USA, 2016. Association for Computing Machinery.
\newblock ISBN 9781450342322.
\newblock \doi{10.1145/2939672.2939778}.
\newblock URL \url{https://doi.org/10.1145/2939672.2939778}.

\bibitem[Ribeiro et~al.(2018)Ribeiro, Singh, and Guestrin]{Ribeiro_Singh_Guestrin_2018}
Marco~Tulio Ribeiro, Sameer Singh, and Carlos Guestrin.
\newblock Anchors: High-precision model-agnostic explanations.
\newblock \emph{Proceedings of the AAAI Conference on Artificial Intelligence}, 32\penalty0 (1), Apr. 2018.
\newblock \doi{10.1609/aaai.v32i1.11491}.
\newblock URL \url{https://ojs.aaai.org/index.php/AAAI/article/view/11491}.

\bibitem[Shin et~al.(2018)Shin, Kim, Shin, and Xiao]{privacyenhancedmatfact}
Hyejin Shin, Sungwook Kim, Junbum Shin, and Xiaokui Xiao.
\newblock Privacy enhanced matrix factorization for recommendation with local differential privacy.
\newblock \emph{IEEE Transactions on Knowledge and Data Engineering}, 30\penalty0 (9):\penalty0 1770--1782, 2018.
\newblock \doi{10.1109/TKDE.2018.2805356}.

\bibitem[Toubiana et~al.(2010)Toubiana, Narayanan, Boneh, Nissenbaum, and Barocas]{toubiana2010adnostic}
Vincent Toubiana, Arvind Narayanan, Dan Boneh, Helen Nissenbaum, and Solon Barocas.
\newblock Adnostic: Privacy preserving targeted advertising.
\newblock \emph{NDSS}, 2010.

\bibitem[Vora and Samala(2023)]{vora2023scoring}
Jian Vora and Pranay~Reddy Samala.
\newblock Scoring black-box models for adversarial robustness.
\newblock In \emph{The Second Workshop on New Frontiers in Adversarial Machine Learning}, 2023.
\newblock URL \url{https://openreview.net/forum?id=iy4xRjfdid}.

\bibitem[Warner(1965)]{warner1965randomized}
Stanley~L Warner.
\newblock Randomized response: A survey technique for eliminating evasive answer bias.
\newblock \emph{Journal of the American Statistical Association}, 60\penalty0 (309):\penalty0 63--69, 1965.

\end{thebibliography}

\appendix
\onecolumn

\section{Other candidate posterior sampling algorithms}{\label{sec:candidate_multi_selection}}

In this section, we introduce three alternative posterior sampling algorithms—$\mathcal{A}_{avg}$, $\mathcal{A}_{avg-realuser}$, and $\mathcal{A}_{sat}$ by instantiating Algorithm \ref{alg:alg_gen} with various utility functions and posterior distributions, as outlined in Table \ref{tab:Different-mechanisms}. Notably, two of these algorithms $\mathcal{A}_{avg}$ and $\mathcal{A}_{sat}$ do not limit sampling to the set of user feature vectors. Consequently, we define the posterior distribution $\mathcal{L}^{cap}_{\eta}(f)$ below.

\textbf{Distribution $\mathcal{L}^{cap}_{\eta}(f)$}: A sample $s$ from distribution $\mathcal{L}^{cap}_{\eta}(f)$ is constructed from $f$ by adding independent Laplace noise of parameter $\eta$ to each dimension, then capping each component between $[0,1]$ and uniformly scaling each component such that the sum of feature values corresponding to liked and disliked movies remain unity. We ensure that the sum is unity to maintain consistency with the property of the feature constructed for each user, as described in Section \ref{subsec:feature_engineering}.

In algorithms $\mathcal{A}_{avg}$ and $\mathcal{A}_{avg-realuser}$, the utility function $u_{(s)}(.)$
is defined by averaging over the scores of all top $r$ movies in $B_i$ (formally defined in Equation \eqref{eq:utility_tr2}). Recall that $\mathcal{R}^r_f$ denotes the set of $r$ movies with the highest ratings corresponding to the user feature $f \in \real^d$.


\begin{equation}{\label{eq:utility_tr2}}
    u_{(s)}^{avg,r}(f,B_i) := \sum_{b \in B_i} u_{\mathcal{M}} (f,b) \mathbbm{1}_{b \in \mathcal{R}^r_f}
\end{equation}

Observe that setting $t=\infty$ in the utility function $u^{t,r}_{(s)}(f,.)$ (defined in Equation \eqref{eq:utility_2}) gives us the utility function $u^{avg,r}_{(s)}(f,.)$. 

%

\begin{table}[htbp]
\centering
\begin{tabular}{ |c| c| c| }
\hline
 Algorithm  & Posterior distribution $\mathcal{P}$ & Utility $u_{(s)}(.)$ \\ 
\hline
$\mathcal{A}_{\text{ig-sig}}$ & $\text{Uni}(\{f_a\}_{a\in A^{tr}})$ & $u_{(s)}^{t=1,r=100}(.)$  \\
\hline
$\mathcal{A}_{\text{avg}}$ & $\mathcal{L}^{cap}_{\eta}(f)$ & $u_{(s)}^{avg, r=100}(.)$ \\
\hline
$\mathcal{A}_{\text{avg-realuser}}$ & $\mathcal{L}^{realuser}_{\eta}(f)$ & $u_{(s)}^{avg, r=100}(.)$ \\
\hline
$\mathcal{A}_{\text{sat}}$ &  $\mathcal{L}^{cap}_{\eta}(f)$ & $u_{(s)}^{t=1,r=100}(.)$ \\
\hline
$\mathcal{A}_{\text{sat-realuser}}$ &  $\mathcal{L}^{realuser}_{\eta}(f)$ & $u_{(s)}^{t=1,r=100}(.)$\\
\hline


\end{tabular}

\caption{Instantiation of Algorithm \ref{alg:alg_gen} for utilities $u_{(s)}(.)$ and posterior distribution $\mathcal{P}$}
\label{tab:Different-mechanisms}
\end{table}



\section{Experimental results of candidate multi-selection algorithms}{\label{sec:candidate_simulation}}


Our experimental setup is identical to the setup described in Section \ref{sec:experimental_setup} by uniformly sampling an user uniformly at random 1500 times to calculate the average dis-utility across the experiments. We further split the results into two parts and in the first part, we measure the dis-utility induced by geographic DP by computing function $d_i$ (defined in Equation \ref{eq:disutility_inter}) assuming the user directly receives the set of $k$ movies $B_i$. In the second part, we measure dis-utility induced by frugal model by computing $d_f$ (defined in Equation \ref{eq:disutility_final}) where the agent uses frugal model choose its movie $b_f$.

At a high level, we make the following observations.

\begin{itemize}
    \item The posterior sampling algorithms $\mathcal{A}_{avg-realuser}$ and  $\mathcal{A}_{sat-realuser}$ have much smaller dis-utility than their counterparts which do not restrict to sampling from the set of users in training set.
    \item Further selection of naive utility function $u^{avg,r}(.)$ which averages the utility across all top $r$ movies results in higher dis-utility since it is not commensurate with the fact that the user's agent (privacy delegate) chooses its best movie from $B_i$.
\end{itemize}

\paragraph{Dis-utility due to geographic DP} We measure the dis-utility $d_i$ assuming the user directly receives the set of $k$ movies $B_i$ and has access to $\mathcal{M}$. In table \ref{tab:candidate_no_local_model_disutilities_deepnn}, we show that the algorithms $\mathcal{A}_{{sat}}$ and $\mathcal{A}_{{sat-realuser}}$ have far smaller disutility $d_i$ than the mechanisms $\mathcal{A}_{nopost}(k=1)$ and $\mathcal{A}_{nopost-realuser}(k=1)$ which returns a single movie without multi-selection and posterior sampling. 

\begin{table*}[thbp]
\begin{centering}
\begin{tabular}{|c|c|c|c|c|c|c|c|c|c|}
           \hline
                \multirow{2}{*}{{\centering $\eta$}} & \multirow{2}{*}{\parbox{2cm}{\centering $\mathcal{A}_{nopost} \newline (k=1)$}} & \multirow{2}{*}{\parbox{2.5cm}{\centering $\mathcal{A}_{nopost-realuser}\newline (k=1)$}}&\multicolumn{3}{|c|}{Values of $k$ ($\mathcal{A}_{sat}$)} &\multicolumn{3}{|c|}{Values of $k$($\mathcal{A}_{sat-realuser}$)}  \\
                \cline{4-9} & & & 2 & 3 & 5 & 2 & 3 & 5 \\
                \hline
                0.01 & 0.0447 & 0.1147 &  0.0209 & 0.0129 & 0.0077 & 0.0656 & 0.0506 & 0.0416 \\
\hline
0.03 & 0.1748 & 0.1591 & 0.0881 & 0.0603 & 0.0377 & 0.0644 & 0.0441 & 0.0275\\
\hline
0.05 & 0.2622 & 0.1945 & 0.1341 & 0.0976 & 0.0604 & 0.0811 & 0.0552 & 0.0354\\
\hline
0.1 & 0.3749 & 0.2826 & 0.192 & 0.1384 & 0.092 & 0.1225 & 0.0849 & 0.0476  \\
\hline
0.15 & 0.3806 & 0.3548 & 0.2292 & 0.1695 & 0.1066 & 0.1394 & 0.098 & 0.0577\\
\hline
0.2 & 0.4236 & 0.3897 & 0.2453 & 0.1788 & 0.1221 & 0.1532 & 0.113 & 0.0673 \\
\hline
\end{tabular}
\caption{Dis-utility $d_f$ of algorithms $\mathcal{A}_{{sat}}$ and $\mathcal{A}_{{sat-realuser}}$ against baselines}
\label{tab:candidate_no_local_model_disutilities_deepnn}
\end{centering}
\end{table*}



\begin{figure*}[htbp]
\centering
\begin{minipage}{.5\textwidth}
  \centering
  \includegraphics[width=8cm, height=4cm]{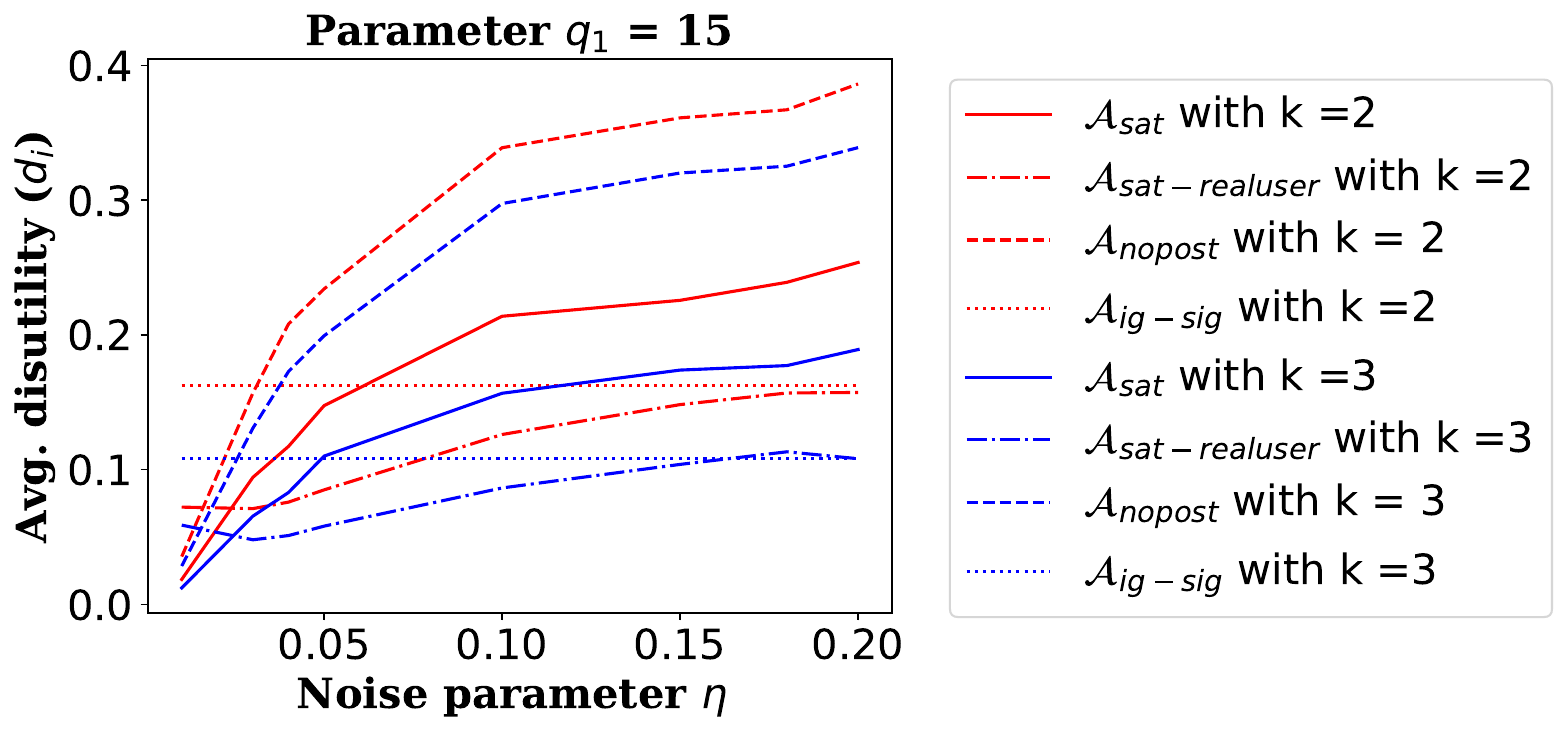}
 
\end{minipage}%
\begin{minipage}{.5\textwidth}
  \centering
  \includegraphics[width=8cm, height=4cm]{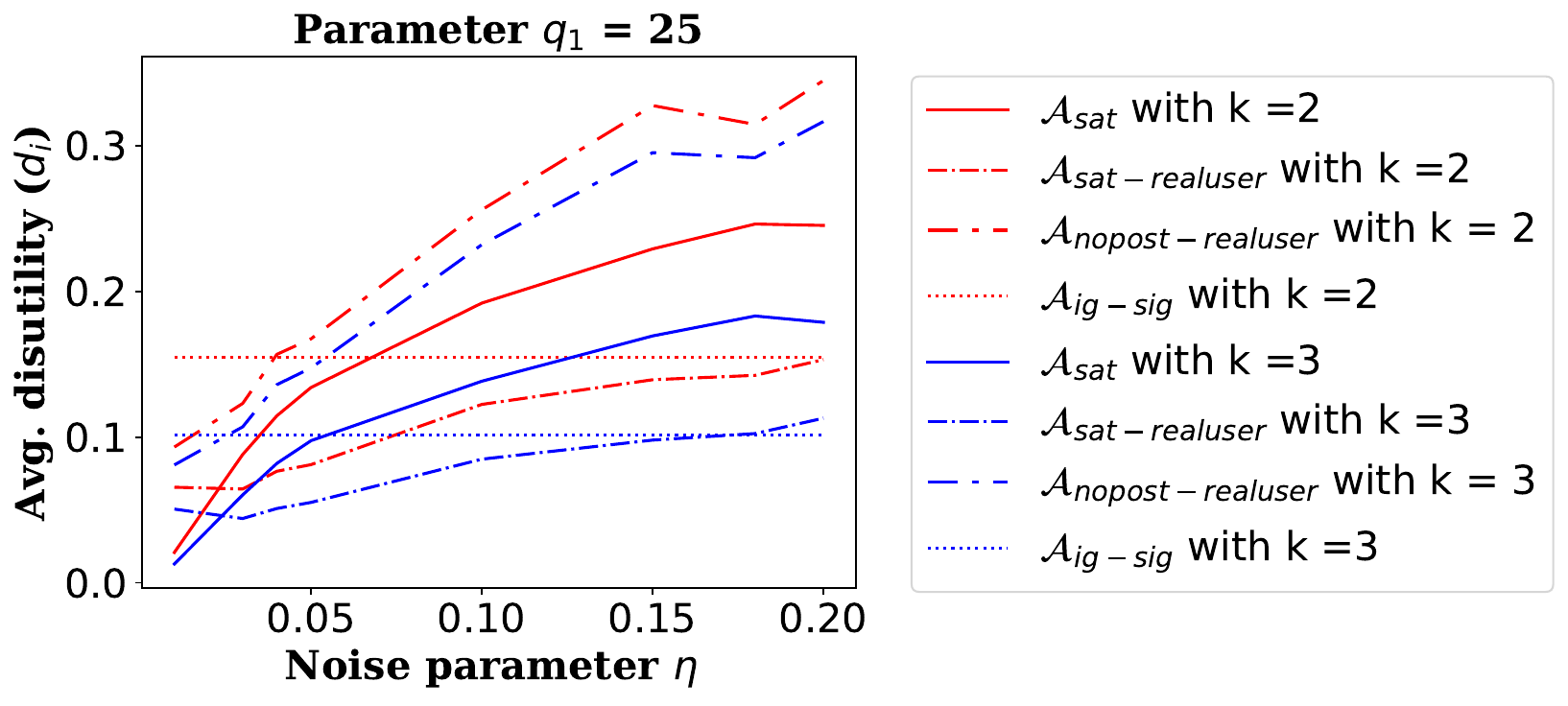}
  
\end{minipage}
\caption{Plotting dis-utility of different algorithms with noise}
  \label{fig:candidate_diff_mech}
\end{figure*}

In Figure \ref{fig:candidate_diff_mech}, we compare various different algorithms $\mathcal{A}_{{sat}}$ and $\mathcal{A}_{{sat-realuser}}$ over the baselines $\mathcal{A}_{nopost-realuser}$ and $\mathcal{A}_{nopost}$ and $\mathcal{A}_{ig-sig}$ for different values of $q_1$. One may observe that the mechanism $\mathcal{A}_{sat-realuser}$ gives the best dis-utility for $\eta$ around $[0.05, 0.2]$. Note that while mechanisms $\mathcal{A}_{sat}$ performs the best for very small values of $\eta$, its performance degrades for moderate values of $\eta$. We conjecture this is because the feature vectors the server samples corresponds to feature vectors of non-existent users and the ground-truth model $\mathcal{M}$ may not give a very good prediction on those feature vectors since it is not trained on them. Further, the mechanism $\mathcal{A}_{nopost-realuser}$ gives the worst dis-utility for most values of $\eta$. Recall that the mechanism $\mathcal{A}_{ig-sig}$ ignores the signal and thus its dis-utility is independent of the noise level $\eta$.

\begin{figure*}[htbp]
\centering
\begin{minipage}{.5\textwidth}
  \centering
  \includegraphics[width=8cm, height=4cm]{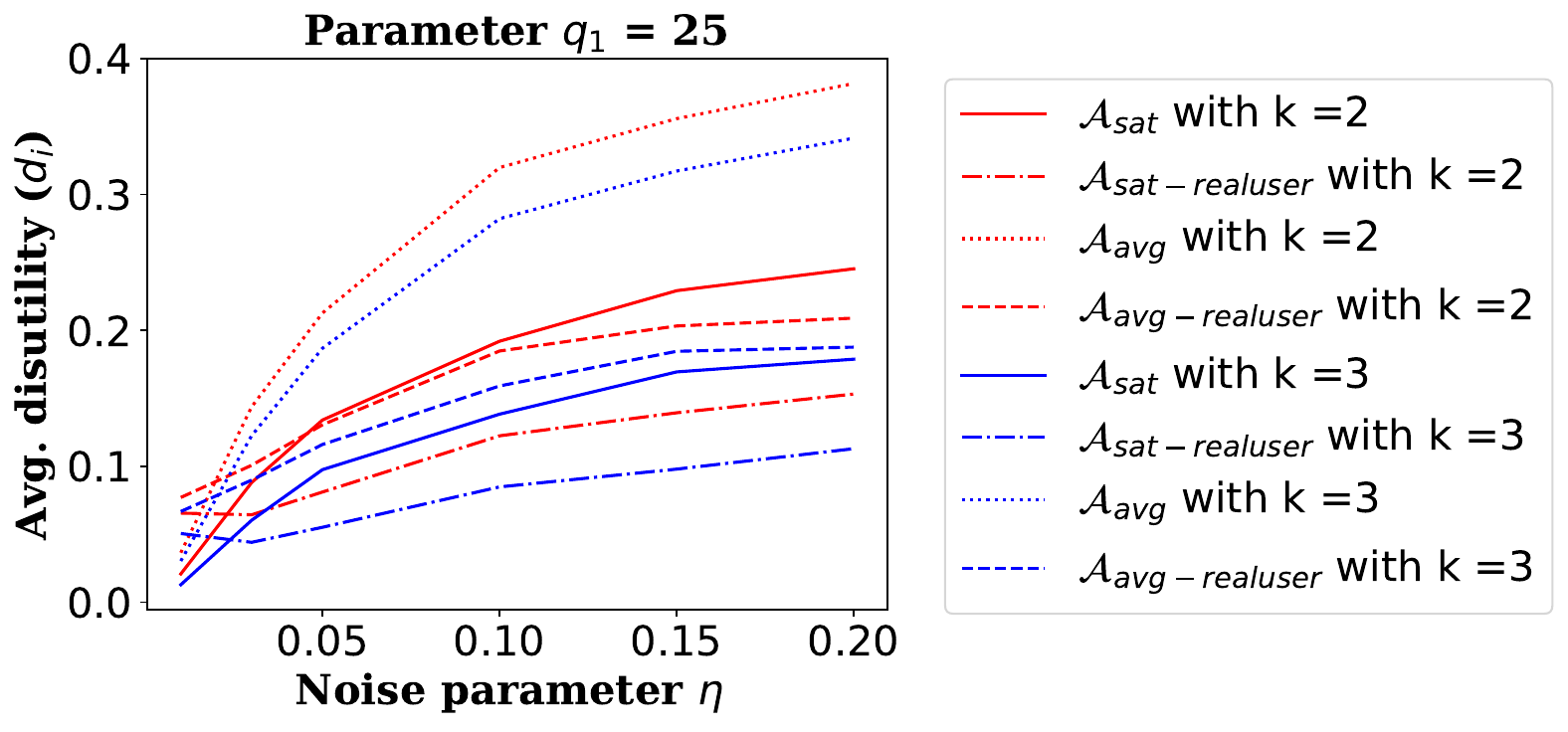}
 
\end{minipage}%
\begin{minipage}{.5\textwidth}
  \centering
  \includegraphics[width=8cm, height=4cm]{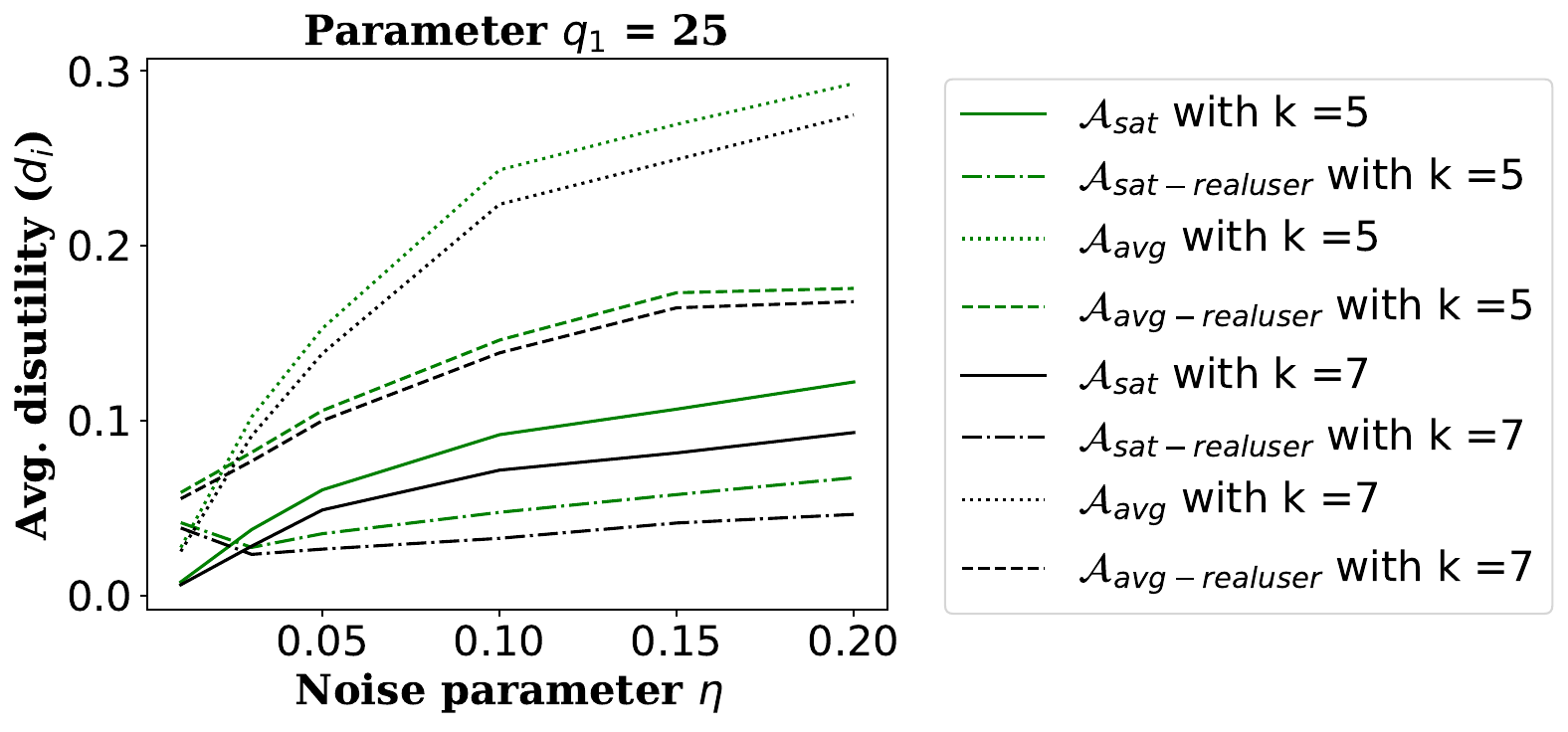}
  
\end{minipage}
\caption{Plotting dis-utility of different algorithms with noise}
  \label{fig:candidate_diff_mech_v2}
\end{figure*}

\newpage
In Figure \ref{fig:candidate_diff_mech_v2}, we compare the algorithms $\mathcal{A}_{sat}$ and $\mathcal{A}_{sat-realuser}$ against the algorithms $\mathcal{A}_{avg}$ and $\mathcal{A}_{avg-realuser}$ (which measure the dis-utility by averaging it over the sampled users). Similar to the observation in Figure \ref{fig:diff_mech}, we can observe that the algorithms $\mathcal{A}_{sat-realuser}$ and $\mathcal{A}_{avg-realuser}$ have lower dis-utility than the other mechanisms possibly because they only sample feature vectors from the true users in $A^{tr}$. However, the algorithm  $\mathcal{A}_{sat-realuser}$ has the lowest dis-utility since the utility function $u_{(s)}^{t=1,r}$ is commensurate with the fact that the user selects its best movie from the set of $k$ sent movies $B_i$ unlike the utility function $u_{(s)}^{avg,r}$ which sums over the utility of all top $r$ movies.  

\begin{figure*}[htbp]
\centering
\begin{minipage}{.5\textwidth}
  \centering
  \includegraphics[width=8cm, height=4cm]{plots_deepnn/deepnn_no_local_sat_k_5_i.pdf}
 
\end{minipage}%
\begin{minipage}{.5\textwidth}
  \centering
  \includegraphics[width=8cm, height=4cm]{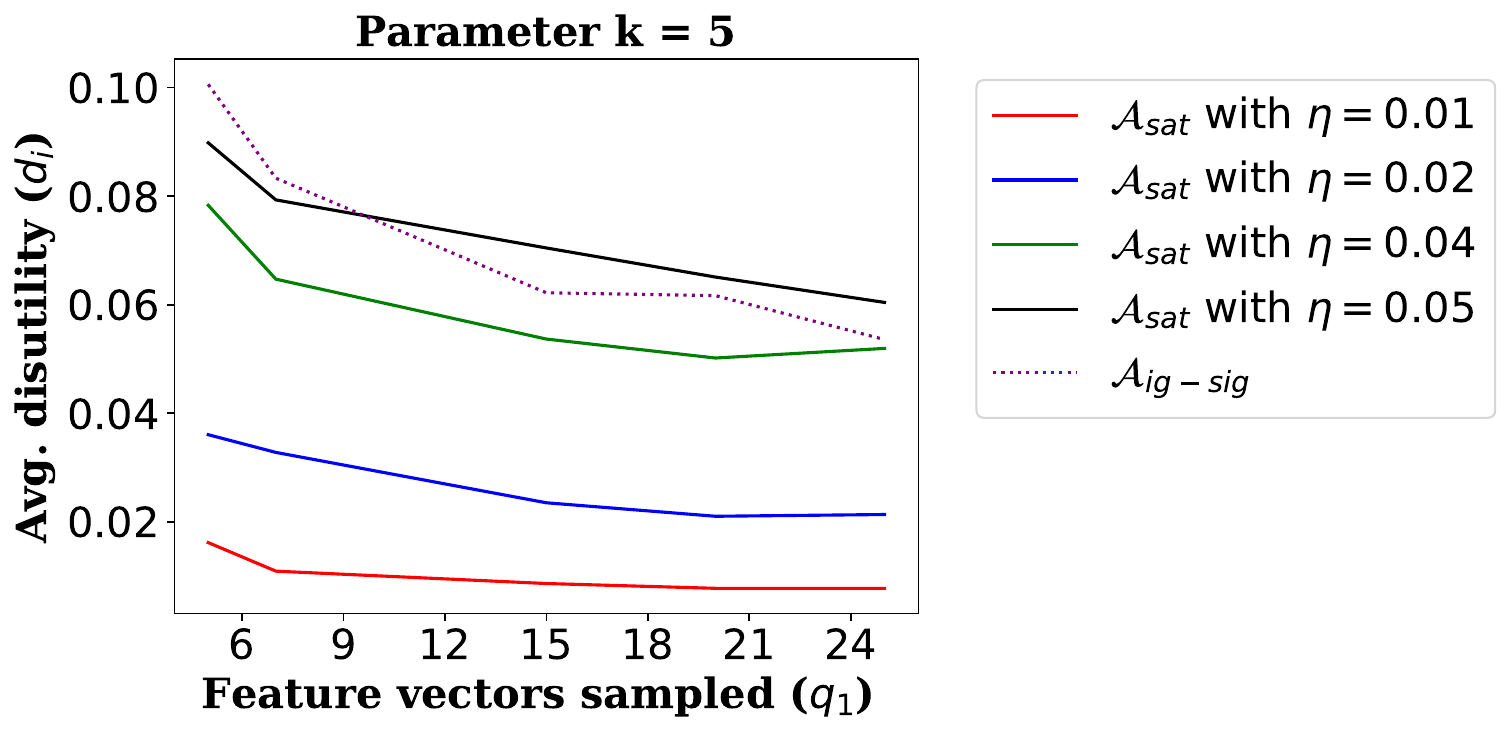}
  
\end{minipage}
\caption{Dis-utility of different algorithms for varying $q_1$}
 \label{fig:candidate_plot_q1}

\end{figure*}

In Figure \ref{fig:candidate_plot_q1}, we compare the mechanisms $\mathcal{A}_{ig-sig}$ and $\mathcal{A}_{sat-realuser}$ and we can observe that the dis-utility monotonically decreases with $q_1$ saturating at $q_1$ around 25.

\begin{figure*}[htbp]
\centering
\begin{minipage}{.5\textwidth}
  \centering
  \includegraphics[width=8cm, height=4cm]{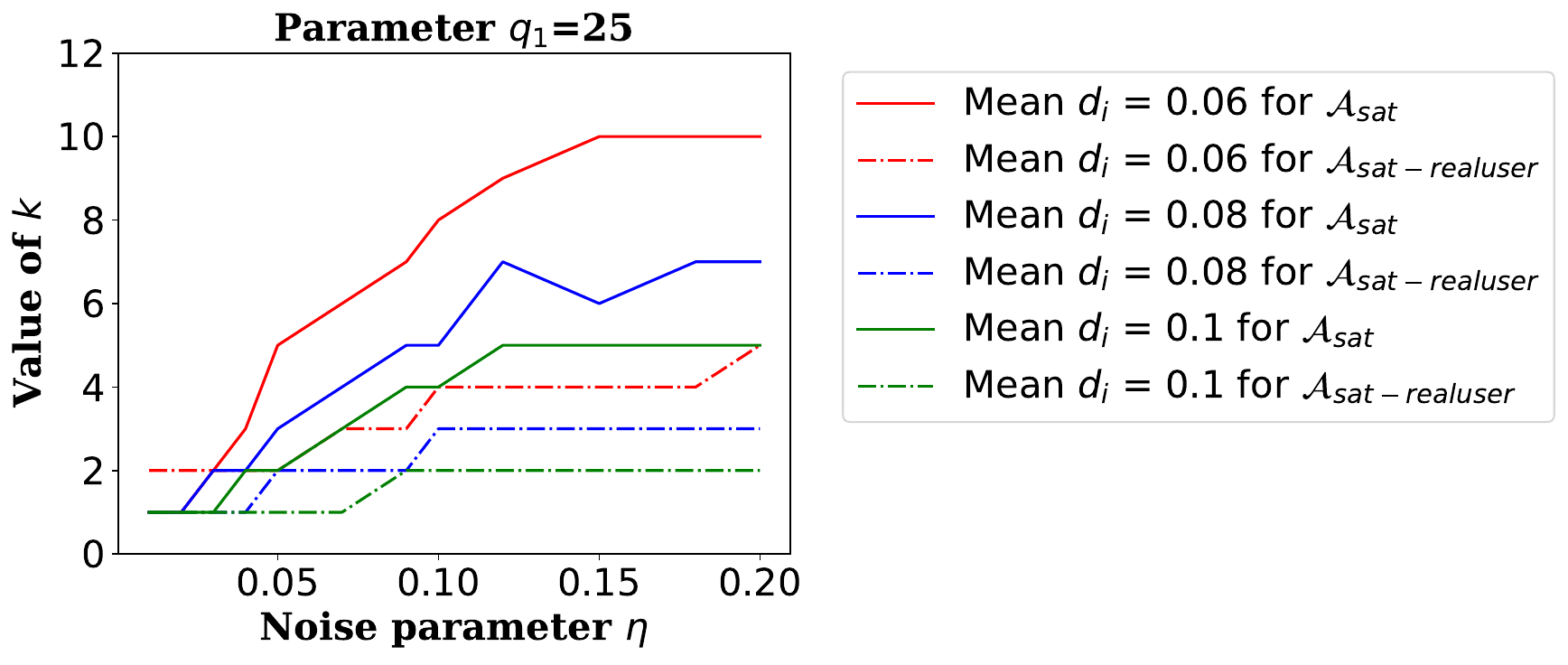}
 
\end{minipage}%
\begin{minipage}{.5\textwidth}
  \centering
  \includegraphics[width=8cm, height=4cm]{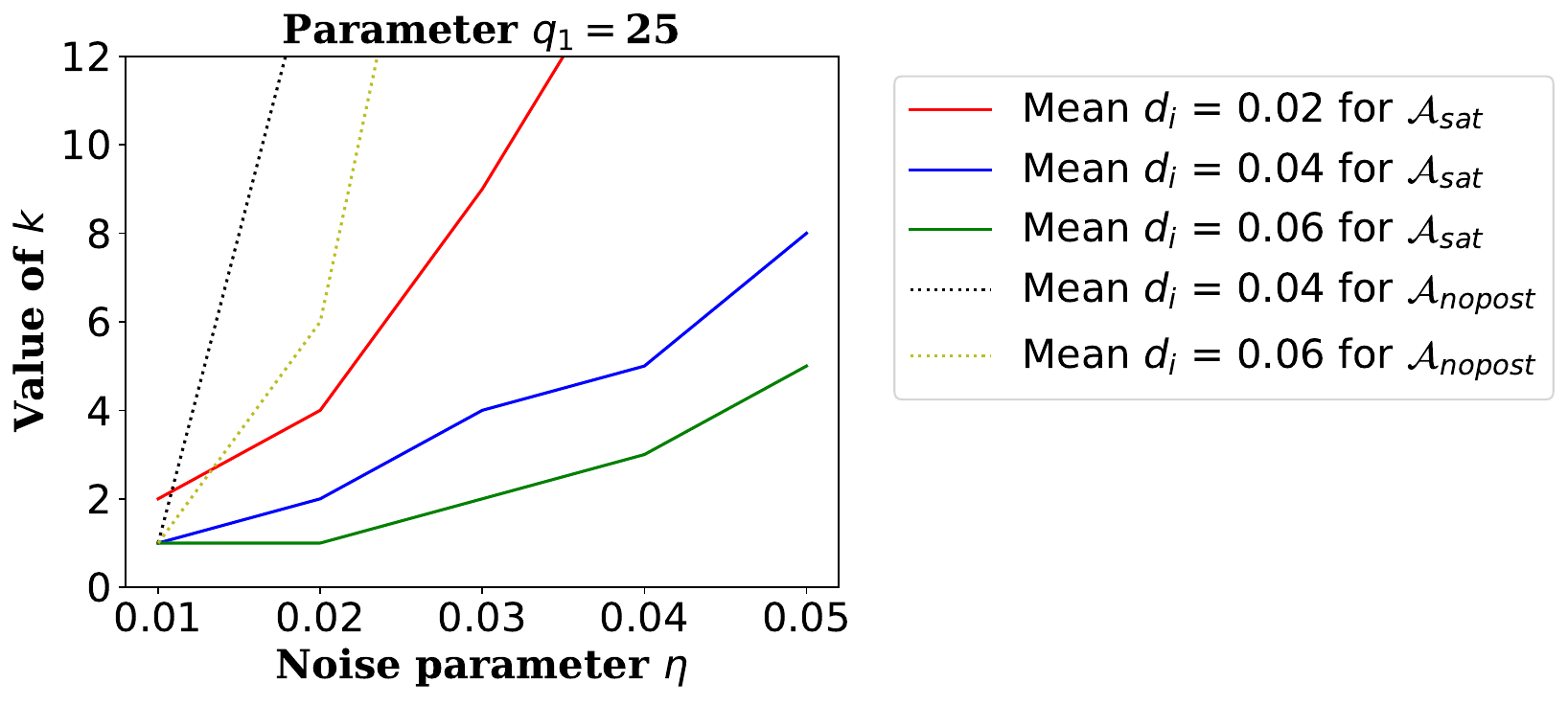}
  
\end{minipage}
\caption{Variation of $k$ with $\eta$ for different values of dis-utility $d_i$}
 \label{fig:candidate_disutility_vs_eta}
\end{figure*}

In Figure \ref{fig:candidate_disutility_vs_eta}, we aim to understand how much does $k$-selection buys us i.e for a fixed level of mean disutility $d_i$ for different $\eta$ by comparing against various algorithms namely $\mathcal{A}_{{sat}}$ and $\mathcal{A}_{sat-realuser}$. Further, even under stringent accuracy constraints (low $d_i$) the value of $k$ goes to atmost 10-12.


\paragraph{Dis-utility due to the geographic DP and frugal model $\mathfrak{m}$} In this part, we measure the dis-utility $d_f$ assuming the user's agent (privacy delegate) uses the frugal model $\mathfrak{m}$ to select the movie $b_f$ from the set of $k$ movies $B_i$. In table \ref{tab:local_model_disutilities_deepnn}, we show that the mechanisms $\mathcal{A}_{{sat}}$ and $\mathcal{A}_{{sat-realuser}}$ has far smaller dis-utility $d_f$ than the mechanisms $\mathcal{A}_{nopost}(k=1)$ and $\mathcal{A}_{nopost-realuser}(k=1)$ which returns a single movie without multi-selection and posterior sampling.

\begin{table*}[htbp]
\begin{centering}
\begin{tabular}{|c|c|c|c|c|c|c|c|c|}
           \hline
                \multirow{2}{*}{{\centering $\eta$}} & \multirow{2}{*}{\parbox{2cm}{\centering $\mathcal{A}_{nopost} \newline (k=1)$}} & \multirow{2}{*}{\parbox{2.5cm}{\centering $\mathcal{A}_{nopost-realuser}\newline (k=1)$}}&\multicolumn{3}{|c|}{Values of $k$ ($\mathcal{A}_{sat}$)} &\multicolumn{3}{|c|}{Values of $k$($\mathcal{A}_{sat-realuser}$)}  \\
                \cline{4-9} & & & 2 & 3 & 5 & 2 & 3 & 5 \\
                \hline
0.03 & 0.194 & 0.159 & 0.124 & 0.113 & 0.124& 0.111 & 0.121 & 0.121 \\
\hline
0.05 & 0.268 & 0.194 & 0.192 & 0.178 & 0.176 & 0.132 & 0.117 & 0.12 \\
\hline
0.1 & 0.385 & 0.283 & 0.279 & 0.252 & 0.256 & 0.169 & 0.15 & 0.143 \\
\hline
0.15 & 0.383 & 0.355 & 0.317 & 0.291 & 0.294 & 0.189 & 0.171 & 0.152 \\
\hline
0.2 & 0.423 & 0.39 & 0.331 & 0.292 & 0.286 & 0.198 & 0.173 & 0.167\\
\hline


\end{tabular}
\caption{Dis-utility $d_f$ under $\mathcal{A}_{sat-realuser}$ and $\mathcal{A}_{sat}$ against baselines}
\label{tab:local_model_disutilities_deepnn2}
\end{centering}
\end{table*}

\begin{figure*}[htbp]
\centering
\begin{minipage}{.5\textwidth}
  \centering
  \includegraphics[width=8cm, height=4cm]{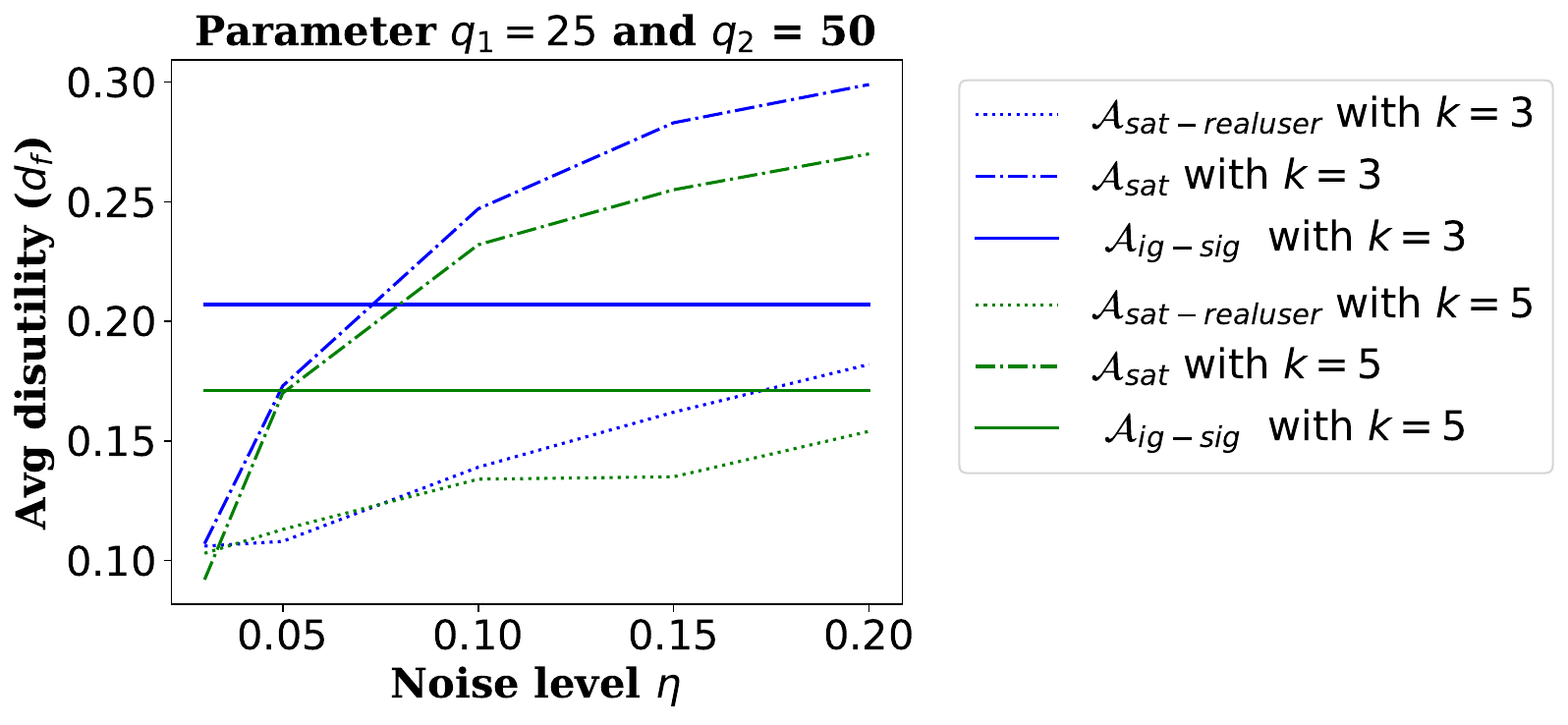}
  
\end{minipage}%
\begin{minipage}{.5\textwidth}
  \centering
  \includegraphics[width=8cm, height=4cm]{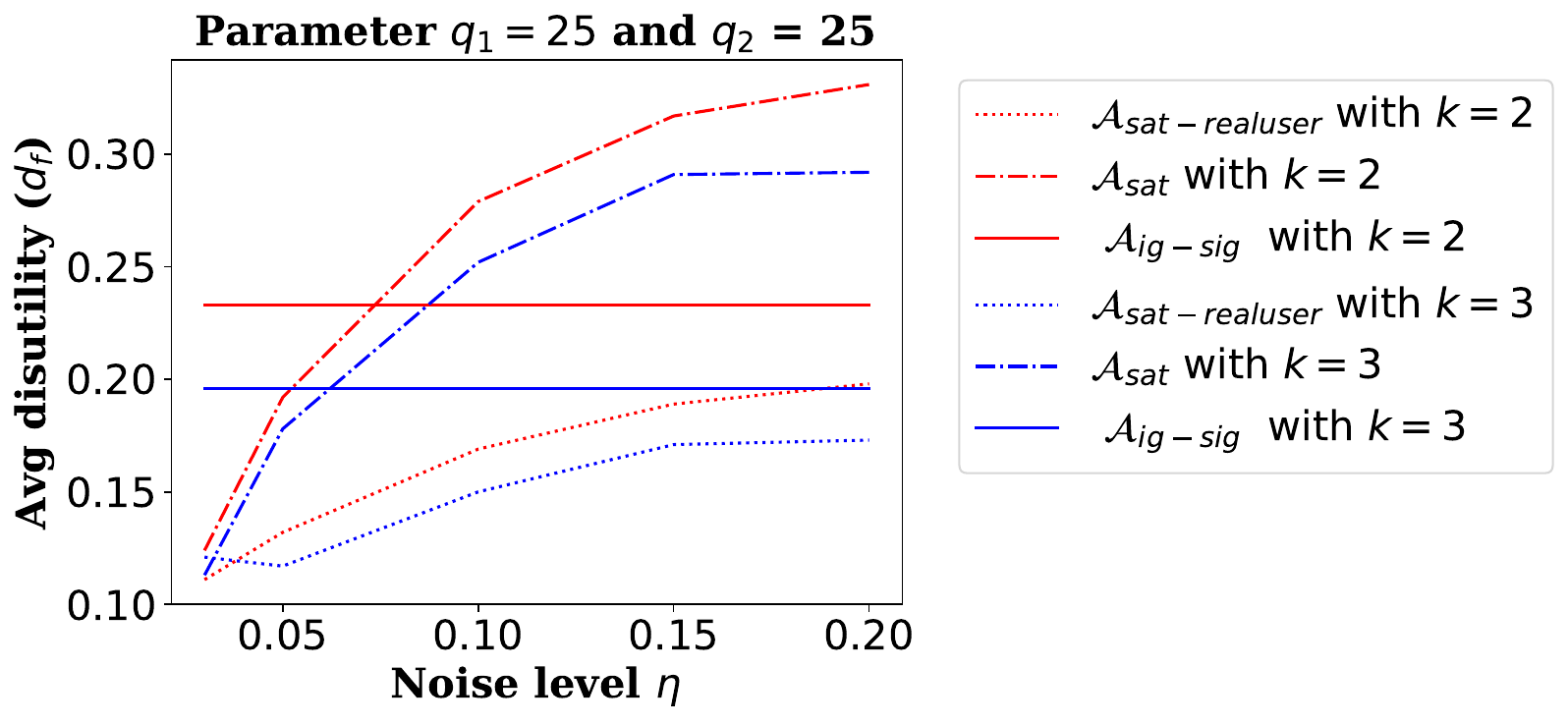}
\end{minipage}
\caption{Disutility $d_f$ of different algorithms}
\label{fig:diff_mech_local_model}
\end{figure*}

Figure \ref{fig:diff_mech_local_model} compares the dis-utilities $d_f$ of the mechanisms $\mathcal{A}_{sat}$ and $\mathcal{A}_{sat-realuser}$ against the baseline mechanism $\mathcal{A}_{ig-sig}$ and shows that the mechanism $\mathcal{A}_{sat-realuser}$ has the smallest dis-utility $d_f$. Similar to the plot in Figure \ref{fig:diff_mech}, we can observe that the mechanism $\mathcal{A}_{sat}$ has higher disutility possibly because the server samples from feature vectors of non-existent users. 

\balance 
\end{document}